\PassOptionsToPackage{dvipsnames}{xcolor} 

\documentclass[pmlr,twocolumn,10pt]{jmlr} 

\usepackage{booktabs}
\usepackage{siunitx}

\usepackage{xcolor}  
\usepackage{lipsum}
\usepackage{lmodern}
\usepackage[breakable,skins]{tcolorbox}
\usepackage{colortbl}
\usepackage{soul}
\usepackage{array}
\usepackage[switch]{lineno}

\newcommand{\repourl}{https://github.com/hyesunyun/MedLitSpin}

\newcommand{\hlblue}[1]{\sethlcolor{Cerulean!50}\hl{#1}} 
\newcommand{\hlorange}[1]{\sethlcolor{Peach!50}\hl{#1}} 

\definecolor{orange}{HTML}{FF8100}

\theorembodyfont{\upshape}
\theoremheaderfont{\scshape}
\theorempostheader{:}
\theoremsep{\newline}

\jmlrvolume{287}
\jmlryear{2025}
\jmlrsubmitted{} 
\jmlrpublished{} 
\jmlrworkshop{Conference on Health, Inference, and Learning (CHIL) 2025}

\title[Do LLMs Fall for Spin in Medical Literature?]{Caught in the Web of Words: \\ Do LLMs Fall for Spin in Medical Literature?}

\author{%
 \Name{Hye Sun Yun} \Email{yun.hy@northeastern.edu}\\
 \addr Northeastern University, Boston, MA, USA\\
 \Name{Karen Y.C. Zhang} \Email{zhang.yuchen@northeastern.edu}\\
 \addr Northeastern University, Boston, MA, USA\\
 \Name{Ramez Kouzy} \Email{rkouzy@mdanderson.org}\\
 \addr The University of Texas MD Anderson Cancer Center, Houston, Texas, USA\\
 \Name{Iain J. Marshall} \Email{iain.marshall@kcl.ac.uk}\\
 \addr King’s College London, London, UK\\
 \Name{Junyi Jessy Li} \Email{jessy@utexas.edu}\\
 \addr The University of Texas at Austin, Austin, Texas, USA\\
 \Name{Byron C. Wallace} \Email{b.wallace@northeastern.edu}\\
 \addr Northeastern University, Boston, MA, USA
}

\begin{document}

\maketitle

\begin{abstract}
Medical research faces well-documented challenges in translating novel treatments into clinical practice. 
Publishing incentives encourage researchers to present ``positive'' findings, even when empirical results are equivocal. 
Consequently, it is well-documented that authors often \emph{spin} study results, especially in article abstracts. 
Such spin can influence clinician interpretation of evidence and may affect patient care decisions.
In this study, we ask whether the interpretation of trial results offered by Large Language Models (LLMs) is similarly affected by spin. 
This is important since LLMs are increasingly being used to trawl through and synthesize published medical evidence. 
We evaluated 22 LLMs and found that they are across the board \emph{more} susceptible to spin than humans. 
They might also propagate spin into their outputs: We find evidence, e.g., that LLMs implicitly incorporate spin into plain language summaries that they generate. 
We also find, however, that LLMs are generally capable of recognizing spin, and can be prompted in a way to mitigate spin's impact on LLM outputs. 
\end{abstract}

\paragraph*{Data and Code Availability}
This paper uses the cancer-related medical abstract dataset \citep{boutron2014impact}, which is available as a Data Supplement. 
All code required to reproduce the experiments in this paper is available at \url{\repourl}.

\paragraph*{Institutional Review Board (IRB)}
This research did not require IRB approval as it is designated as Not Human Subject Research.

\begin{figure}[t]
\floatconts
  {fig:fig1}
  {\caption{Authors of medical articles sometimes \emph{spin} their reporting of trial results. We find that LLMs are susceptible to this when ``reading'' medical abstracts, more so than human experts.}}
  {\includegraphics[width=\linewidth]{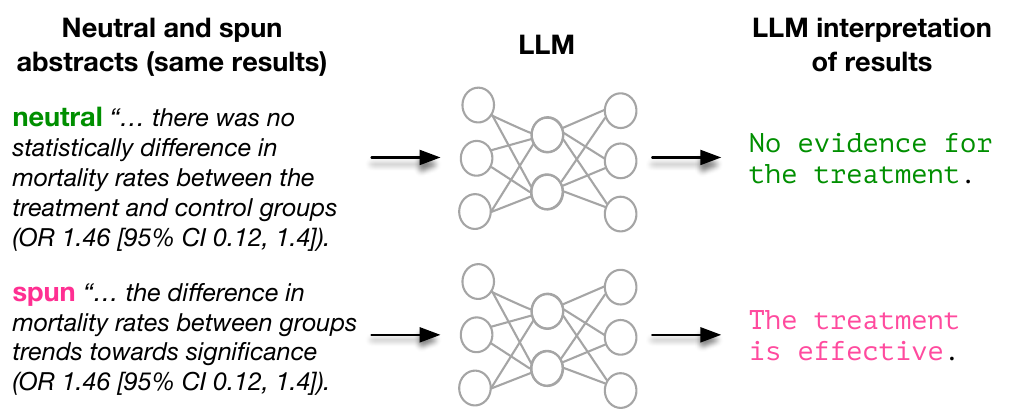}}
\end{figure}

\begin{table*}[t]
\footnotesize
\floatconts
  {tab:example_abstracts}
  {\caption{Example of an abstract with spin \citep{osborne2002double} and an abstract rewritten without spin. The highlighted portions of text show the changes between the original and the rewritten version.}}
  {\begin{tabular}{p{0.48\textwidth} p{0.48\textwidth}}
  \toprule
  \multicolumn{1}{c}{\bfseries Abstract with Spin} & \multicolumn{1}{c}{\bfseries Abstract Rewritten without Spin}\\
  \midrule
  
  \rowcolor{gray!20}
  \scriptsize\textbf{Purpose}: To compare the efficacy and tolerability of treatment A with comparator B in the treatment of advanced breast cancer in patients whose disease progresses on prior endocrine treatment.
 & \scriptsize\textbf{Purpose}: To compare the efficacy and tolerability of treatment A with comparator B in the treatment of advanced breast cancer in patients whose disease progresses on prior endocrine treatment.
\\
  \scriptsize\textbf{Patients and methods}: In this double‐blind, double‐dummy, parallel‐group study, postmenopausal patients were randomized to receive either treatment A or comparator B. The primary end point was time to progression (TTP). Secondary end points included objective response (OR) rate, duration of response (DOR), and tolerability. & \scriptsize\textbf{Patients and methods}: In this double‐blind, double‐dummy, parallel‐group study, postmenopausal patients were randomized to receive either treatment A or comparator B. The primary end point was time to progression (TTP). Secondary end points included \hlblue{time to treatment failure (TTF),} objective response (OR) rate, duration of response (DOR), and tolerability.
\\
  \rowcolor{gray!20}\scriptsize\textbf{Results}: Patients (n = 400) were followed for a median period of 16.8 months. Treatment A was \hlorange{as effective as} comparator B in terms of TTP (hazard ratio, 0.92; 95\% confidence interval [CI], 0.74 to 1.14; P =.43); median TTP was 5.4 months with treatment A and 3.4 months with comparator B. \hlorange{OR rates were 17.5\% with both treatments. Clinical benefit rates (complete response + partial response + stable disease $>$ or = 24 weeks) were 42.2\% for treatment A and 36.1\% for comparator B (95\% CI, ‐4.00\% to 16.41\%; P =.26). In responding patients, median DOR (from randomization to progression) was 19.0 months for treatment A and 10.8 months for comparator B. Using all patients,} DOR was significantly greater for treatment A compared with comparator B; the ratio of average response durations was 1.35 (95\% CI, 1.10 to 1.67; P $<$ 0.01). Both treatments were well tolerated.
 & \scriptsize\textbf{Results}: Patients (n = 400) were followed for a median period of 16.8 months. Treatment A was \hlblue{not more effective than} comparator B in terms of TTP (hazard ratio, 0.92; 95\% confidence interval [CI], 0.74 to 1.14; P =.43); median TTP was 5.4 months with treatment A and 3.4 months with comparator B. \hlblue{There was no statistically significant difference between the 2 groups for TTF. Median TTF was 4.6 months for treatment A and 3.3 months for comparator B (HR, 0.96; 95\% CI, 0.77 to 1.19; P = .69). At the time of this data analysis, the rate of deaths was respectively for treatment A and comparator B, 35.4\% (n=73) vs. 33.5\% (n=65). OR rates were 17.5\% with both treatments.} DOR was statistically significantly greater for treatment A compared with comparator B; the ratio of average response durations was 1.35 (95\% CI, 1.10 to 1.67; P $<$ 0.01). Both treatments were well tolerated.
\\
  \scriptsize\textbf{Conclusion}: Treatment A was \hlorange{at least as effective as comparator B, with efficacy end points slightly favoring treatment A. Treatment A represents an additional treatment option} for postmenopausal women with advanced breast cancer whose disease progresses on tamoxifen therapy. & \scriptsize\textbf{Conclusion}: Treatment A was \hlblue{not more effective than comparator B} for postmenopausal women with advanced breast cancer whose disease progresses on tamoxifen therapy.\\
  \bottomrule
  \end{tabular}}
\end{table*}

\section{Introduction}
\label{section:intro}
Randomized controlled trials (RCTs) form the cornerstone of evidence-based medicine. 
Healthcare providers often base clinical decisions on trial findings, mostly as presented in article abstracts \citep{christakis2000physicians, hopewell2008consort, berwanger2009quality, marcelo2013comparison}. 
Unlike full texts, abstracts are concise and easily accessible, unlike full articles which are often behind paywalls \citep{smith2017knowledge, piwowar2018state, day2020open}. Clinicians are 2.4 times more likely to read an abstract than a full article \citep{islamaj2009understanding}.

The information presented in abstracts alone is sufficient to alter most clinicians' care decisions \citep{barry2001family}. 
But while convenient, relying on abstracts brings drawbacks: Several studies have shown that biomedical abstracts may not present an entirely accurate impression of the underlying study results \citep{boutron2010reporting, chiu2017spin, wayant2019evaluation, jellison2020evaluation, nowlin2022spin}. 
Abstracts frequently present overly optimistic interpretations of results, reflecting broader publication pressures that favor positive findings. \citep{begg1988publication, dickersin1990existence, easterbrook1991publication, dickersin1993publication, devito2019catalogue}.

\emph{Spin} refers to reporting strategies that overstate the benefits of experimental treatments beyond what is supported by empirical evidence.
For instance, spin might seek to distract readers from statistically nonsignificant results, overstate the efficacy, and/or understate the harms of a treatment.
This is also known as ``hype'' \citep{sumner2016exaggerations, yavchitz2016new, prasad2020malignant}.
About 60\% of abstracts of a sample of RCTs indexed in PubMed with statistically nonsignificant results for the primary outcome contain spin \citep{boutron2010reporting}. 

Elsewhere, \citet{boutron2014impact} showed that spin in cancer RCT abstracts can affect the expert clinician and medical researcher interpretation of trial results, such that they are more likely to rate treatments as beneficial despite the fact that the primary outcome measure is statistically nonsignificant. 
This implies that spin in medical literature abstracts may impact patient treatment. 

Large Language Models (LLMs) are increasingly being used to process medical literature, and will increasingly mediate the consumption of published evidence via automated synthesis, summarization, and simplification \citep{shaib-etal-2023-summarizing, goldsack2025leveraging}.
In this work, we ask: \textbf{How susceptible are LLMs to spin in medical articles?} 
To answer this question empirically, we evaluated 22 LLMs on three tasks: \textbf{(1)} \textit{How well can LLMs detect the presence of spin in abstracts of RCT reports?} As LLMs are increasingly used to aid human decision-making in medicine \citep{thirunavukarasu2023large}, their ability to detect spin and interpret results accordingly will be key. \textbf{(2)} \textit{How do LLMs interpret the same trial results when presented with spun versus unspun abstracts?} 
LLMs may be capable of recognizing spin when explicitly asked to do so, and yet still be susceptible to it insofar as their interpretation of study results may be influenced by spin; we aim to quantify this. 
\textbf{(3)} \textit{To what extent might LLMs propagate or amplify spin in medical abstracts when generating simplified versions?} One practical use of LLMs in medicine is \emph{simplification} to enable lay consumption of new technical content in medicine \citep{devaraj2022evaluating}. We investigate how the degree of spin in technical abstracts affects automatically generated plain language summaries (PLS). 

Our findings reveal that, at least in oncology, LLMs are more susceptible to spin in medical abstracts than clinicians and medical researchers. This susceptibility persists despite their ability to accurately detect spin when explicitly prompted. Even though LLMs recognize spin, they still propagate it to downstream tasks, such as generating simplified versions of technical abstracts. These findings demonstrate LLMs' vulnerability to spin in medical literature, emphasizing the importance of careful implementation when using them to process clinical research. To address this issue, we also explored simple mitigation strategies that may help reduce the impact of spin on LLMs.

\begin{figure*}[t]
\floatconts
  {fig:detection_accuracy}
  {\caption{Spin detection task accuracies for all LLMs. The average accuracy of all models was 0.67 (solid red vertical line), well above the random baseline (gray dashed vertical line). That said, this plot shows considerable variance across models with respect to their spin detection capabilities.}}
  {\includegraphics[width=0.8\linewidth]{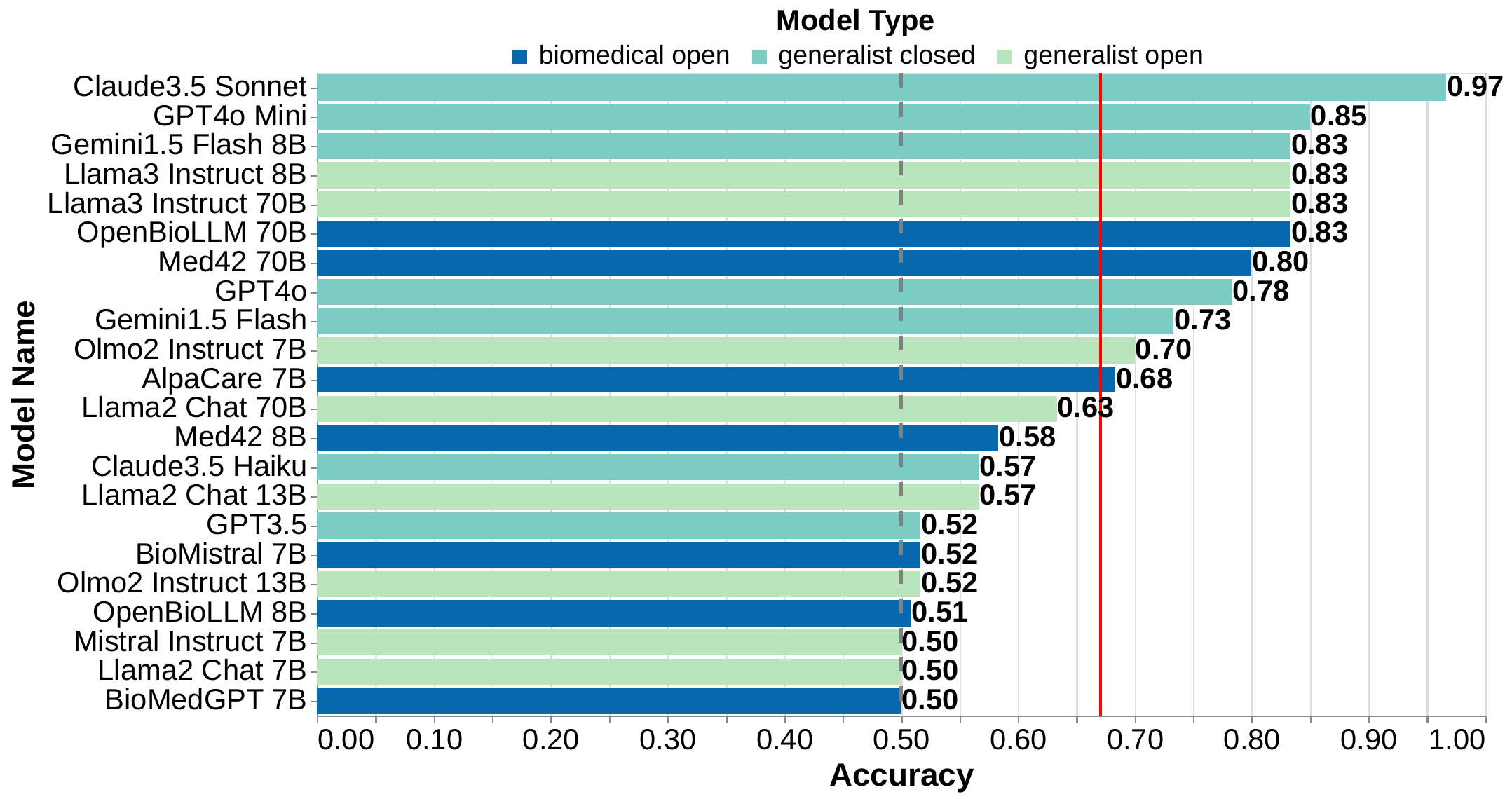}}
\end{figure*}

\section{Experimental Setup}
\label{section:setup}

To empirically evaluate LLM susceptibility to spin in medical abstracts, we ran three experiments measuring how different LLM outputs are when given abstracts with and without spin. 
We base these analyses on data manually collected in prior work on spin in oncology and its influence on medical experts such as clinicians and principal investigators in cancer research (\citealt{boutron2014impact}; Section \ref{section:data}).

In the first experiment, we measured whether LLMs can spot spin in medical abstracts. 
The other two experiments examined downstream effects of any susceptibility to spin: First, we assessed whether LLMs ``interpret'' results differently when the underlying abstract has been spun, and then we measured if any such bias is propagated to automatically simplified abstracts.

\subsection{Data}
\label{section:data}
In all of our experiments we used a small, manually curated dataset from \citet{boutron2014impact}. 
This comprises 60 abstracts (both real and synthetic) paired to compare results reported with and without spin.
The 30 source abstracts are from cancer-related RCTs that met two criteria: (1) reported statistically nonsignificant results for all primary outcomes of the studied intervention, and (2) contained language in their results and conclusion sections that overstated the benefit of the intervention.\footnote{Two independent evaluators, specifically trained to identify ``spin'', evaluated all selected abstracts. Any discrepancies between reviewers were discussed until they reached a consensus \citep{boutron2014impact}.}
All of the underlying trials comprised two arms (intervention and control) and included at least 100 patients.

Researchers in this prior work \citep{boutron2014impact} manually edited each of these abstracts to remove spin, yielding 30 ``neutral'' matched abstracts.\footnote{This was done following guidelines outlined in the main text of \citet{boutron2014impact}.}
This process controlled for length: ``neutralized'' versions were constrained to be within 25 words of the original (spun) versions. 
Author names, references, journal names, registration numbers, trial names or acronyms, and article titles were removed from both versions of the abstracts. 
Further, treatment descriptions were systematically masked with generic terms (e.g., treatment A and comparator B) to avoid any pre-existing bias individuals may have based on treatment names. Table~\ref{tab:example_abstracts} shows an example of an abstract with spin and a rewritten abstract without.

\subsection{Evaluation}

\paragraph{Spin detection} Using the abstracts from \citet{boutron2014impact}, we prompted LLMs to answer whether or not a given abstract contains spin. 
This is a binary classification problem with a balanced distribution, so we use accuracy as a metric. The exact prompt we used is available in Appendix~\ref{appendix:detection_prompt}.

\begin{figure*}[t]
\floatconts
  {fig:interpretation}
  {\caption{Average mean differences of scores from LLMs for all 5 interpretation questions compared to human experts. Error bars indicate 95\% confidence intervals. A positive mean difference indicates that LLMs interpreted the spun abstract as showing more favorable treatment results while the negative mean difference indicates unspun abstracts to be more favorable. This plot suggests that LLMs, in general, erroneously infer larger differences in results between spun and unspun abstracts than do human experts. \figureref{fig:spin_label_regression_benefit} explains the effect of spin for each LLM.}}
  {\includegraphics[width=\linewidth]{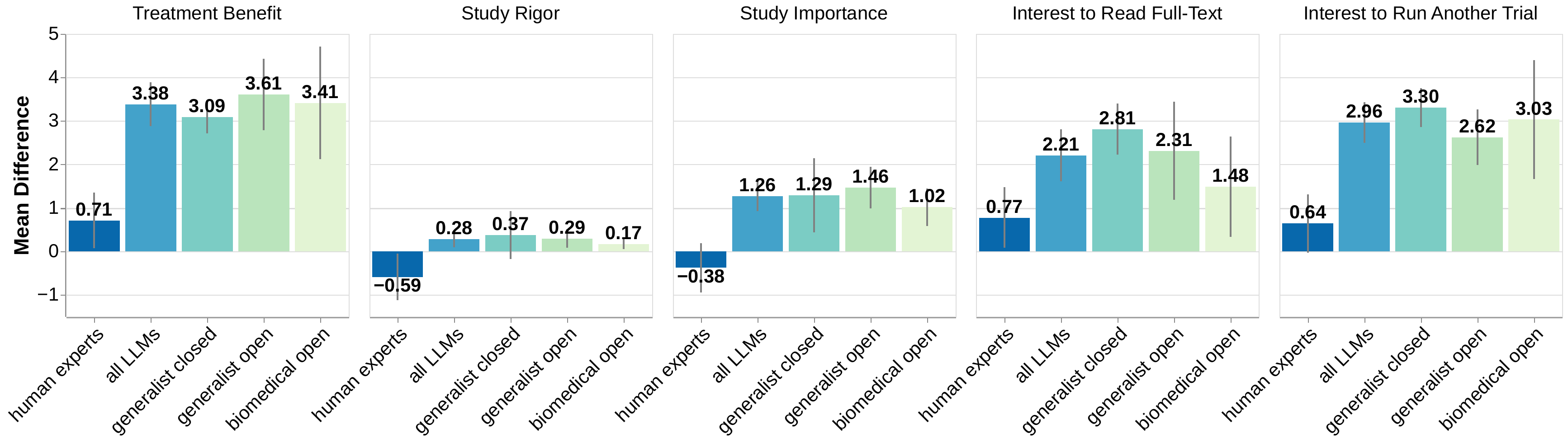}}
\end{figure*}

\paragraph{Interpretation of trial results for spun and unspun abstracts} 
For each abstract, we prompted LLMs to answer the following 5 questions individually on a scale of 0-10: 

\begin{tcolorbox}[colback=blue!5!white,colframe=blue!75!black,title=Interpretation Questions, label=box:questions]
    (1) Based on this abstract, do you think treatment \emph{A} would be beneficial to patients? [very unlikely to very likely]\\
    
    (2) Rate the overall rigor of the study methodology. [not at all rigorous to very rigorous]\\
    
    (3) Rate the importance of the study. [not at all important to very important]\\
    
    (4) Are you interested in reading the full-text article for the study described in the abstract? [not at all interested to very interested]\\
    
    (5) Do you think it would be interesting to run another trial evaluating this treatment? [not interesting at all to very interesting]
\end{tcolorbox}
\noindent These 5 questions were used by \citet{boutron2014impact} for their human experts (i.e., clinicians, authors of RCT papers, and principal investigators in cancer research) study with spun and unspun abstracts. The prompt is provided in Appendix~\ref{appendix:interpretation_prompts}

For each question, we calculated the mean difference in LLM scores between paired abstracts with and without spin. A positive mean difference indicates that LLMs interpreted the spun abstract as presenting more favorable treatment results. We then compared these LLM-based interpretations to previously reported human expert assessments from \citet{boutron2014impact}.

\paragraph{\textbf{Simplifying spun and unspun abstracts}} To examine whether LLMs propagate spin to downstream information for lay people, we analyzed the relationship between spun technical abstracts and their automatically generated plain language versions. 
Ensuring the accessibility of medical literature is crucial, as the average American reads at a 7th- to 8th-grade level, and 36\% of the population has basic or below basic health literacy \citep{kutner2006health, marchand2017readability}. 
LLMs have shown promise as tools for simplifying medical texts into plain language \citep{august2023paper, shaib-etal-2023-summarizing, jeblick2024chatgpt, picton2025assessing}, but if spin influences these outputs, it could mislead lay readers by distorting their interpretation of medical evidence \citep{boutron2019three}.

Using the same set of abstracts as before, we prompted the LLMs to generate plain language summaries suitable for a 5th-grader. 
This aligns with recommendations from the National Institutes of Health (NIH), American Medical Association (AMA), and the U.S. Department of Health and Human Services (HHS) to produce patient education materials at a 5th- to 6th-grade reading level \citep{weiss2007health, us2009simply, brega2015ahrq}.
The full prompt is provided in Appendix~\ref{appendix:simplification_prompt}.
To ensure brevity, we limit simplified summaries to at most 300 tokens. 
We did this at inference time by specifying the maximum number of tokens to generate.

\begin{figure*}
\floatconts
  {fig:spin_label_regression_benefit}
  {\caption{Coefficients from linear regression models with 95\% CI for each LLM showing how much different LLMs overestimate the treatment effects (benefit of treatment), when abstracts contain `spin'. In comparison with human experts (0.71), all LLMs were more susceptible to spin. AlpaCare 7B and Olmo2 Instruct 13B were the most susceptible to spin than others.}}
  {\includegraphics[width=0.8\linewidth]{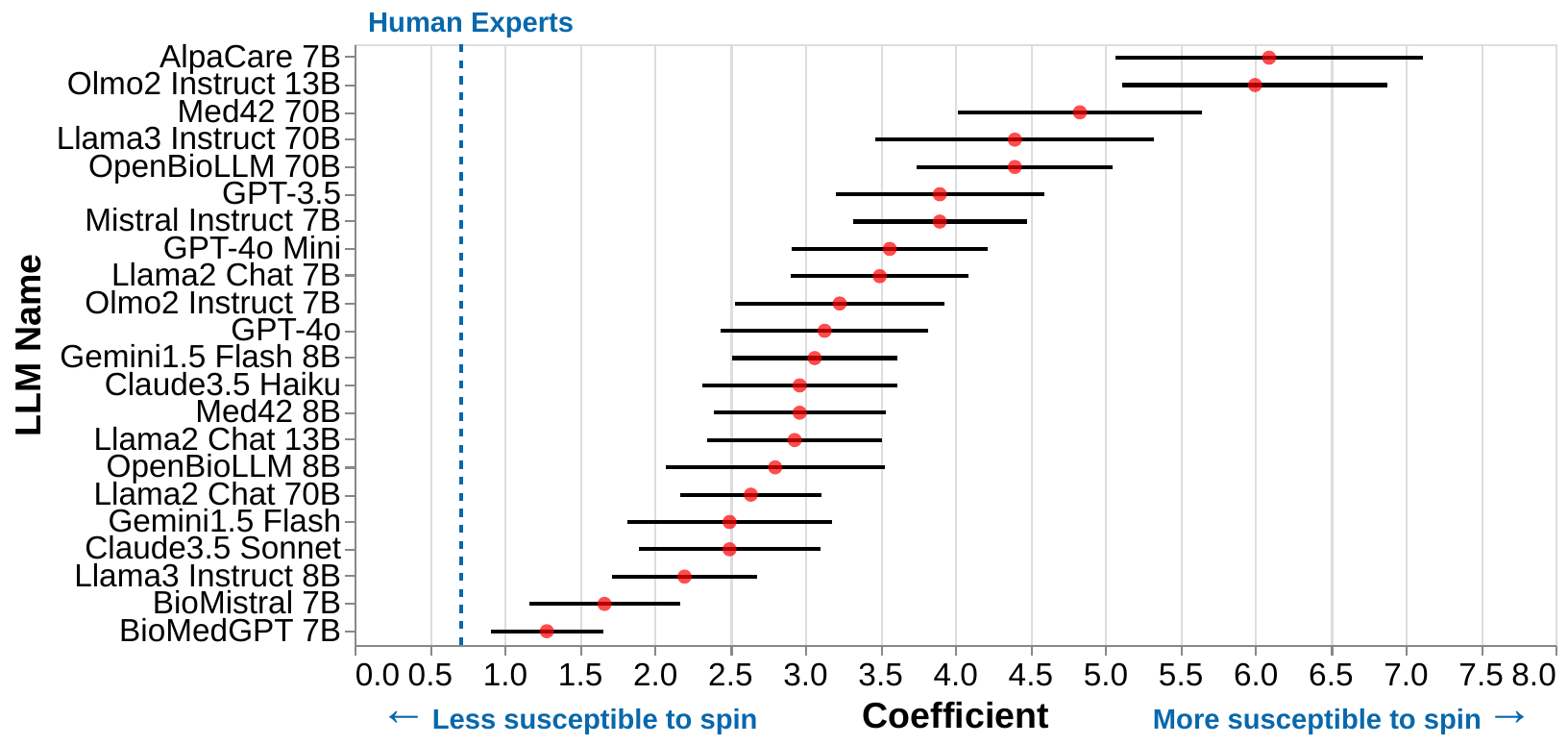}}
\end{figure*}

\subsection{Models}
\label{section:models}

We include in our analysis a variety of LLMs (22 in all), including both open and closed (proprietary) models.
We also include both general and specialized (biomedical) LLMs, and models spanning a range of parameter counts. 

\begin{itemize}
    \item \textbf{Generalist closed-source/proprietary:} Anthropic's Claude 3.5 Haiku and Sonnet, Google's Gemini 1.5 Flash and Flash 8B \citep{team2024gemini}, OpenAI's GPT-3.5 175B \citep{openai2025gpt35turbo}, GPT-4o \citep{hurst2024gpt}, GPT-4o Mini \citep{openai20244o}.
    \item \textbf{Generalist open-weight:} Llama2 7B, 13B, and 70B Chat \citep{touvron2023llama}; Llama3 8B and 70B Instruct \citep{meta2024introducing}; Mistral 7B Instruct v0.1 \citep{jiang2023mistral}; OLMo2 7B and 13B Instruct \citep{olmo20242}.
    \item \textbf{Biomedical open-weight:} Alpacare 7B \citep{zhang2023alpacare}, BioMedGPT 7B \citep{luo2023biomedgpt}, BioMistral 7B \citep{labrak2024biomistral}, Med42-v2 8B and 70B \citep{christophe2024med42}, OpenBioLLM 8B and 70B \citep{pal2024openbiollms}.
\end{itemize}

\section{Results}
\label{section:results}

\paragraph{Detecting spin in abstracts} \figureref{fig:detection_accuracy} reports the accuracies of all LLMs on the spin detection task. 
The average accuracy across all models is 0.67. 
Claude 3.5 Sonnet achieved the highest accuracy score (0.97), followed by GPT-4o Mini (0.85). 
LLMs with accuracy scores close to 0.50 predicted just one label (spin or no spin) for all abstracts. The mean accuracy of generalist closed-source/proprietary LLMs was 0.75. 
Generalist open-weight LLMs had an average accuracy of 0.64, while specialist biomedical open-weight LLMs were comparable (0.63). 
Overall, these results indicate that LLMs are in general moderately to strongly capable of detecting spin. 

\paragraph{Interpretation of (spun) trial results} 
Does spin change LLM interpretation of the reported results? 
The mean differences of the 5 interpretation questions asked to the LLMs are available in \figureref{fig:interpretation}. 
Differences across all interpretation measures from LLMs were generally (much) higher than those observed for human experts.
This suggests that LLMs erroneously infer larger differences in results between spun and unspun abstracts than do human experts.  Additional analysis (Appendix~\ref{appendix:select_models}) confirmed that this held true even for LLMs with spin detection accuracy greater than 0.80. Although human experts found the trials from spun abstracts to be less rigorous or important than from unspun abstracts, they rated trials in abstracts with spin to be more beneficial and were generally more interested in reading the full-text article.

We next quantify the strength of association between the presence of spin and LLM interpretation of results. Specifically, we ran a linear regression with an indicator for the presence of spin in an abstract as predictor, to measure how strongly this correlates with LLM interpretation result scores---in particular, the treatment of benefit measure. 
For simplicity, we ran separate regressions for each LLM, as follows:
\begin{multline*}
    \text{LLM $k$ output score} =
    \beta_{0k} + \\
    \beta_{1k} \cdot (\text{presence or absence of spin})
\end{multline*}
Where $\beta_{0k}$ is an intercept (for LLM $k$), and $\beta_{1k}$ is a coefficient for LLM $k$. 

The coefficients and 95\% confidence intervals for all LLMs for the regression benefit question (Text Box in Section~\ref{box:questions}) are reported in \figureref{fig:spin_label_regression_benefit}. 
This shows that all LLMs considered appear to be more susceptible than human experts to spin in medical abstracts. It also shows that, regardless of spin detection accuracy, the level of susceptibility varies significantly depending on the LLM used.
We report regression results for the other interpretation questions in Appendix~\ref{appendix:linear_models}.

During our analysis, we noticed that a couple of the smaller models such as AlpaCare 7B and BioMedGPT 7B sometimes failed to output valid results for some of the abstracts and questions. We considered model outputs as invalid if they did not output any numerical values between 0 and 10. We removed these inferred outputs from the analysis.

\paragraph{Simplifying spun abstracts}
\label{para:simplifying_spun_abstracts}

We prompted all 22 models (Section~\ref{section:models}) to generate simplified abstracts. The average token length across all simplified abstracts was 208.10 (SD = 67.01).
Selected examples are provided in Appendix~\ref{appendix:simplification_examples}. 
To assess whether LLMs propagate spin when simplifying abstracts into plain language versions, we used three LLMs---GPT-4o Mini, Claude 3.5 Sonnet, and OpenBioLLM 70B---to evaluate the generated summaries. These models were chosen for their strong spin detection performance (\figureref{fig:detection_accuracy}). Using the automatically generated \emph{simplified} versions of abstracts, they answered the same interpretation questions from the Text Box in Section~\ref{section:setup}.

Analysis of LLM-generated plain language summaries showed that spin from the original abstracts persisted in these simplified versions, potentially influencing LLM interpretation of trial results. \figureref{fig:pls_evaluation} shows the analysis of plain language summaries revealing larger score differences between spun and unspun versions, particularly for assessments of treatment benefit, interest in full-text review, and motivation to conduct additional trials. Mean difference scores for plain language summaries varied substantially by evaluator LLM rather than by which LLM generated the summary. Each evaluator model produced consistent mean difference scores across summaries from different LLMs, demonstrating that spin in the original abstracts consistently led to more favorable interpretations of plain language versions.

For treatment benefit assessment, mean differences in LLM-generated summaries were 2.47 for Claude 3.5 Sonnet, 3.59 for GPT-4o Mini, and 4.32 for OpenBioLLM 70B, closely matching the differences observed in evaluations of the original abstracts (\figureref{fig:interpretation} and \ref{fig:spin_label_regression_benefit}). The mean difference scores for rigor and importance of studies were on average lower than the other questions, which also aligns with results from our experiments with the abstracts and the human expert study by \citet{boutron2014impact}.

\begin{figure}[t]
\floatconts
  {fig:pls_evaluation}
  {\caption{Average mean differences of scores from Claude 3.5 Sonnet, GPT-4o Mini, and OpenBioLLM 70B interpreting simplified versions of abstracts with and without spin generated by 22 LLMs. The error bars indicate 95\% confidence intervals. This plot shows that simplified spun abstracts generated by LLMs also exhibit spin.}}
  {\includegraphics[width=\linewidth]{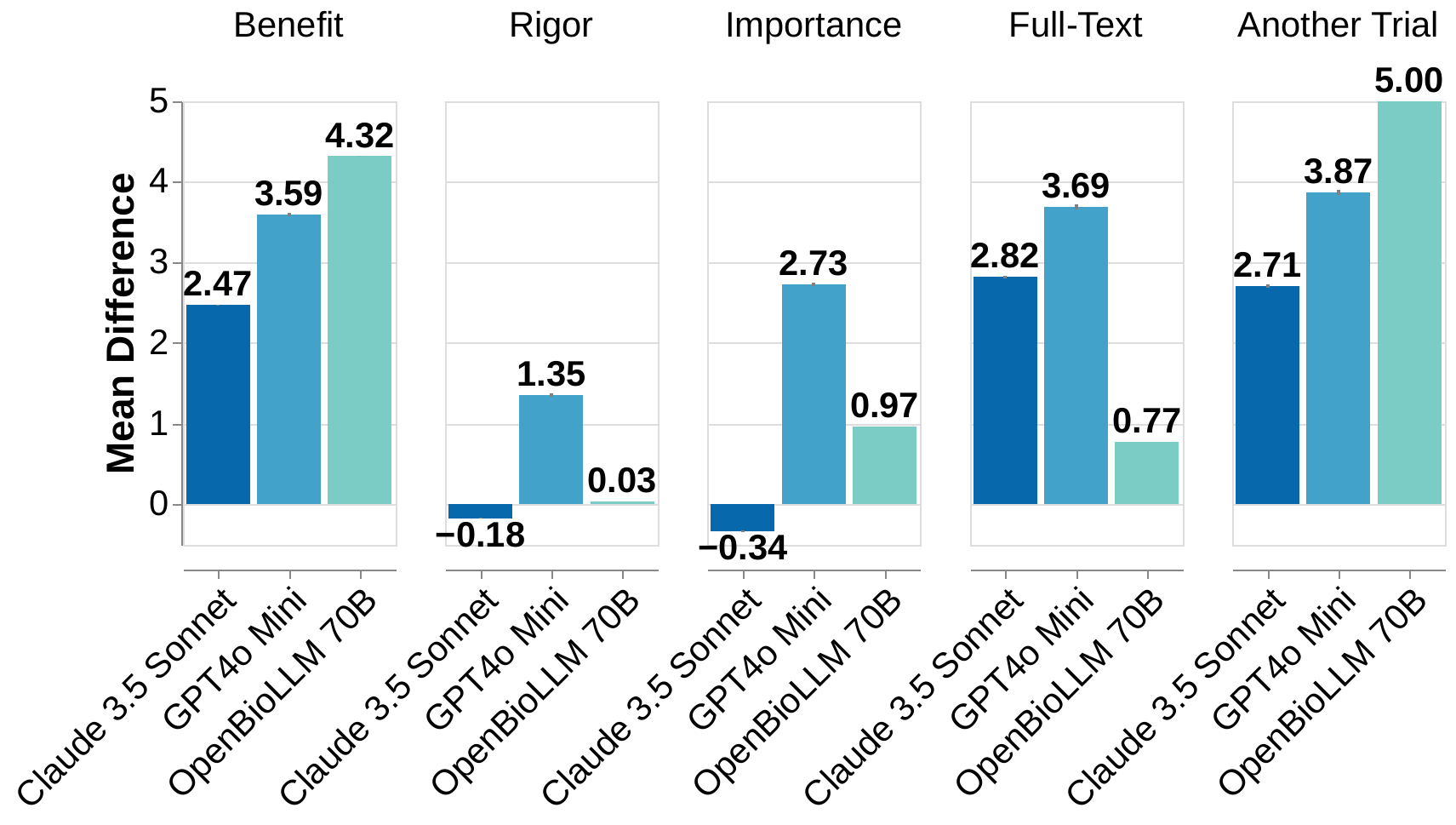}}
\end{figure}

\begin{figure*}[htbp]
\floatconts
  {fig:interpretation_mitigation_comparisons}
  {\caption{Average mean differences of scores across all LLMs using different prompting strategies for 5 interpretation questions compared to human experts. The error bars indicate 95\% confidence intervals. This plot shows that mitigation strategies such as adding additional information on the presence or absence of spin or jointly prompting the model to detect and then interpret can reduce the effect of over-inflating the benefits of the trial results.}}
{\includegraphics[width=\linewidth]{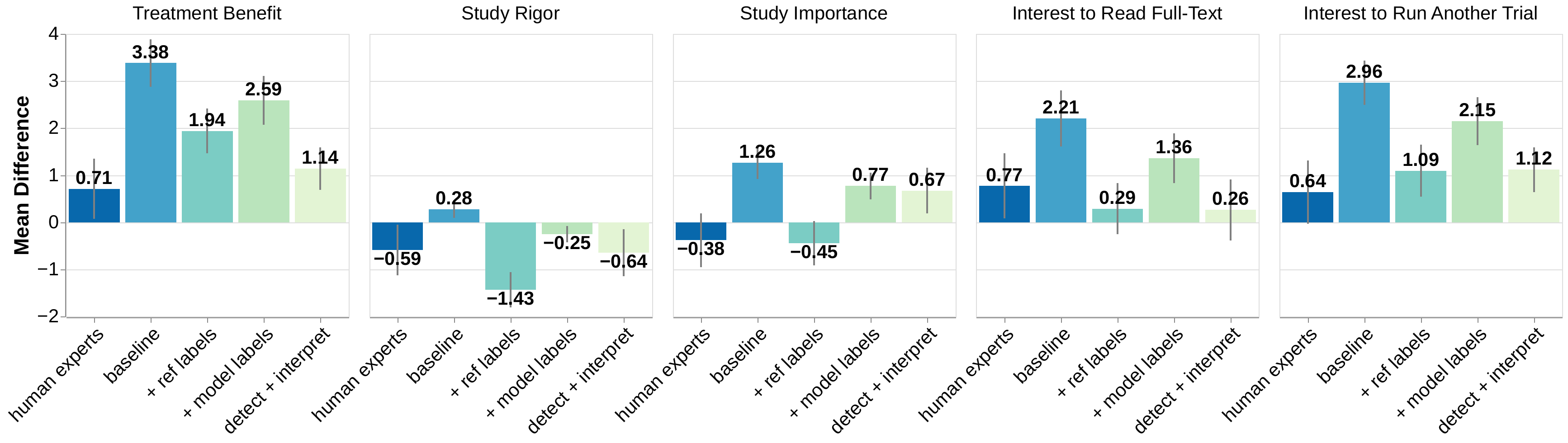}}
\end{figure*}

\section{Reducing the Effect of Spin}

Our findings demonstrate that LLMs are susceptible to spin in medical articles when used without safeguards. To address this vulnerability, we evaluated strategies that might reduce LLMs' tendency to overstate favorable trial results. 

We used two approaches to spin mitigation.
The first approach incorporated explicit spin labels into the interpretation prompts. We tested this strategy using both reference labels and LLM-generated spin assessments, where each model used its own spin detection outputs as labels. Appendix~\ref{appendix:reduce_effect_prompts} provides the exact wording of the prompt.

The second approach we consider involves prompting the LLM to perform both the tasks of spin detection and result interpretation (``joint prompting''). In one prompt, we ask the model to first detect if a given abstract contains spin or not and then to answer the interpretation question based on its answer to the previous question. This combined prompting was done to see if chain-of-thought style prompting \citep{wei2022chain} would allow the model to critically reason for the interpretation task and to be less susceptible to the presence of spin. The prompt for this can also be found in Appendix~\ref{appendix:reduce_effect_prompts}.

\figureref{fig:interpretation_mitigation_comparisons} shows the comparison of our strategies compared to human experts and original prompting (``baseline'') for interpreting trial results. Across all questions we see a decrease in the mean differences when the prompt includes whether a given abstract contains spin or not. 
Providing the reference labels significantly reduces the mean difference, compared to the baseline prompting approach. 
For the treatment benefit question, baseline prompting resulted in an average mean difference of 3.38 across all LLMs.
Including the reference spin label moderated this, yielding a difference of 1.94. 
But even including \emph{inferred} spin labels improves performance to a 2.59 difference.

We also see a significant decrease in the mean differences across all questions when we prompt the LLM to first detect the presence or absence of spin in a given abstract and then answer the interpretation question. The mean differences were much lower than baseline for all questions and were even lower than the reference label approach for treatment benefit, study rigor, and interest to read the full-text. The mean difference for the treatment benefit question was 1.14 compared to 3.38 for the baseline and 1.94 for the reference label approach. Table~\ref{tab:llm_output_example} presents an example of GPT-4o Mini's numerical outputs to the treatment benefit question, using different prompting strategies.

\begin{table*}[ht]
\footnotesize
\floatconts
  {tab:llm_output_example}
  {\caption{This table presents how GPT-4o Mini's numerical score outputs (on a scale of 0 to 10) for the treatment benefit question (Text Box in Section~\ref{box:questions}) vary based on the input abstract and the prompting strategy used. Higher scores indicate a greater perceived treatment benefit. The results highlight differences in scoring between abstracts with and without spin. Notably, the “Detect + Interpret” strategy yields the smallest difference between these abstracts, suggesting that GPT-4o Mini does not interpret the spun abstract as substantially more favorable than the unspun version.}}
  {\begin{tabular}{m{0.29\textwidth} 
      >{\centering\arraybackslash}m{0.11\textwidth} 
      >{\centering\arraybackslash}m{0.11\textwidth} 
      >{\centering\arraybackslash}m{0.11\textwidth} 
      >{\centering\arraybackslash}m{0.11\textwidth} 
      >{\centering\arraybackslash}m{0.11\textwidth}}
  \toprule
      \bfseries Abstract & 
      \bfseries Type & 
      \bfseries Baseline & 
      \bfseries + Ref \newline Label & 
      \bfseries + Model \newline Label & 
      \bfseries Detect \newline + Interpret \\
  \midrule
  
  \rowcolor{gray!20}
  \scriptsize ... The true‐positive (TP) rates were 14.4\% vs. 11.4\% (p=0.035, one‐sided) for the combined colposcopy and intervention A arm compared to colposcopy‐only arm, respectively, in women with either an atypical squamous cell (ASC) or low‐grade squamous intraepithelial lesion (LSIL) cytology result. TP rates were similar between the two arms among women referred for the evaluation of HSIL (high‐grade squamous intraepithelial lesion) ... Conclusion: \hlorange{Combining intervention A with colposcopy provides a clinically meaningful increase in the detection of CIN 2, 3} in women referred for the evaluation of mildly abnormal cytology results. 
  & with spin & 7 & 6 & 7 & 4 \\

  \scriptsize ... There was no statistically significant difference in the true positive rate between the 2 groups. The true positive rate was 19.9\% (218/1096) in the colposcopy‐only arm vs 21.8\% (238/1090) in the colposcopy plus intervention A arm; p=0.143 ... Conclusion: \hlblue{Combining intervention A with colposcopy was not more effective than colposcopy alone for the detection of CIN 2, 3} in women referred for the evaluation of mildly abnormal cytology results. 
  & without spin & 3 & 2 & 2 & 2 \\

  \bottomrule
  \end{tabular}}
\end{table*}

\section{Experiments with More Datasets}
\label{section:additional_dataset}

\subsection{Data}

Our main evaluation relied heavily on a small (though high-quality) dataset of manually selected and edited oncology trials. Here we aim to generalize our findings using a semi-synthetic approach to `spinning' abstracts from a range of medical specialties. This required identifying real-world RCT abstracts containing spin. We sourced three separate datasets from orthopaedics,\footnote{\url{https://osf.io/v63he/overview}}, emergency medicine and obesity,\footnote{\url{https://osf.io/eu5t2/overview}} each consisting of abstracts that had been manually identified as containing spin. 
These datasets comprise 36, 213, and 45 abstracts respectively, totaling 295 abstracts. These abstracts contained spin spanning three categories: (1) Spin due to selective reporting; (2) Spin in the abstract title or results; and (3) Spin in the abstract conclusion. 
To align with our original evaluation data, we included only RCTs exhibiting spin in the abstract title, results, or conclusions, yielding 150 abstracts in total.

Next, we (automatically) generated ``unspun'' versions of each abstract to create paired inputs. 
Specifically, we prompted GPT-4o \citep{hurst2024gpt} to produce an unspun version of each abstract, using the same instructions from \citet{boutron2014impact} that were originally provided to human experts. 
Resulting unspun versions occasionally exhibited formatting inconsistencies relative to the original abstracts, e.g., bolded or uppercased section headers appearing in one version but not the other. These discrepancies were manually corrected for to ensure uniform formatting across each spun–unspun pair, while leaving the substantive content of the abstracts unchanged.

\subsection{Evaluation}

Using our semi-synthetic dataset generated from the process described above, we conducted the same experiments as before, replacing the original input data with our curated abstracts. As no human evaluation was performed, we did not compare model outputs against those of human experts. We report the results below. 

\paragraph{Detecting spin in abstracts}

\begin{figure*}[t]
\floatconts
  {fig:additional_detection_accuracy}
  {\caption{Spin detection task accuracies for all LLMs with the semi-synthetic data. The average accuracy of all models was 0.60 (solid red vertical line), above the random baseline (gray dashed vertical line) but lower than our results from the main evaluation (0.67). This plot shows considerable variance across models with respect to their spin detection capabilities.}}
  {\includegraphics[width=0.8\linewidth]{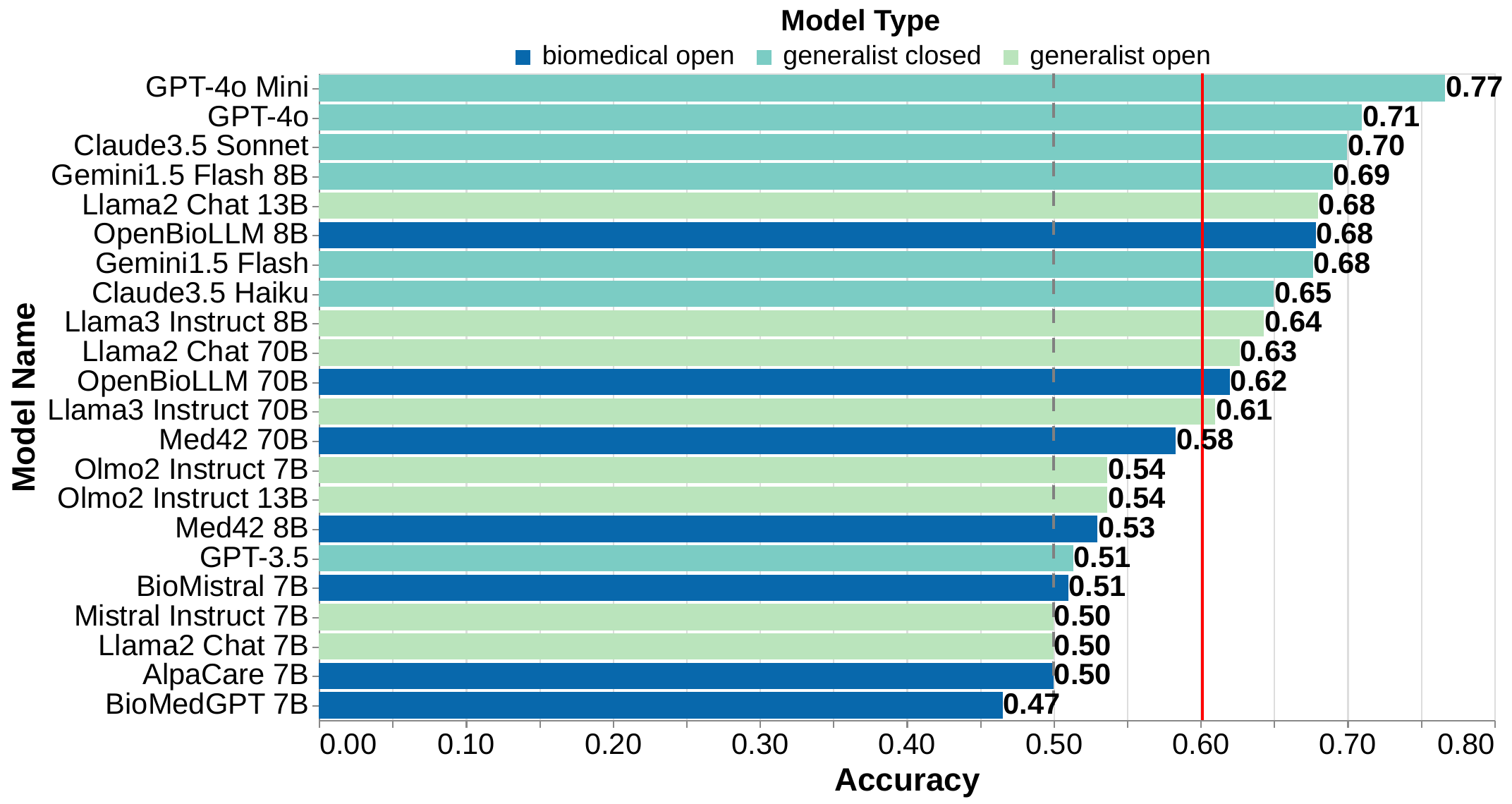}}
\end{figure*}

\figureref{fig:additional_detection_accuracy} reports the spin detection accuracy of all LLMs on the semi-synthetic data. The average accuracy across all models is 0.60, lower than the 0.76 observed on the original high-quality data. GPT-4o Mini achieved the highest accuracy (0.77), followed by GPT-4o (0.71).\footnote{Note that it is possible this owes to our use of GPT-4o to automatically ``unspin'' abstracts in this semi-synthetic setup.} Proprietary generalist LLMs averaged 0.67 overall. Generalist open-weight LLMs had an average accuracy of 0.58, and specialist biomedical open-weight LLMs averaged 0.56. Taken together, these results suggest that LLMs perform above chance on the spin detection task, though performance is noticeably lower on the semi-synthetic data than on the original evaluation set. 

\paragraph{Interpretation of (spun) trial results} 

The mean differences across the five interpretation questions asked to the LLMs are shown in \figureref{fig:additional_interpretation}. Differences across all interpretation measures were generally positive, indicating that LLMs interpreted the spun abstract as reflecting more favorable treatment results (a negative mean difference indicates that the model interpreted the unspun abstract as more favorable for the treatment). However, the overall mean difference scores were considerably smaller than those from the original experiments, likely because the abstracts were generated by a model rather than a human expert.

To quantify the strength of association between the presence of spin and LLM interpretation of results, we fit a linear regression using spin presence as a binary predictor of the perceived treatment benefit measure. The resulting coefficients and 95\% confidence intervals for all LLMs are reported in \figureref{fig:additional_spin_label_regression_benefit}. These results indicate that the majority of LLMs examined are susceptible to spin in medical abstracts to some degree, and that the level of susceptibility varies considerably across models regardless of their spin detection accuracy. This all agrees with the above findings from the main evaluation.

\begin{figure*}[t]
\floatconts
  {fig:additional_interpretation}
  {\caption{Average mean differences of scores from LLMs for all 5 interpretation questions from semi-synthetic data. Error bars indicate 95\% confidence intervals. A positive mean difference indicates that LLMs interpreted the spun abstract as showing more favorable treatment results while the negative mean difference indicates unspun abstracts to be more favorable. \figureref{fig:additional_spin_label_regression_benefit} explains the effect of spin for each LLM.}}
  {\includegraphics[width=\linewidth]{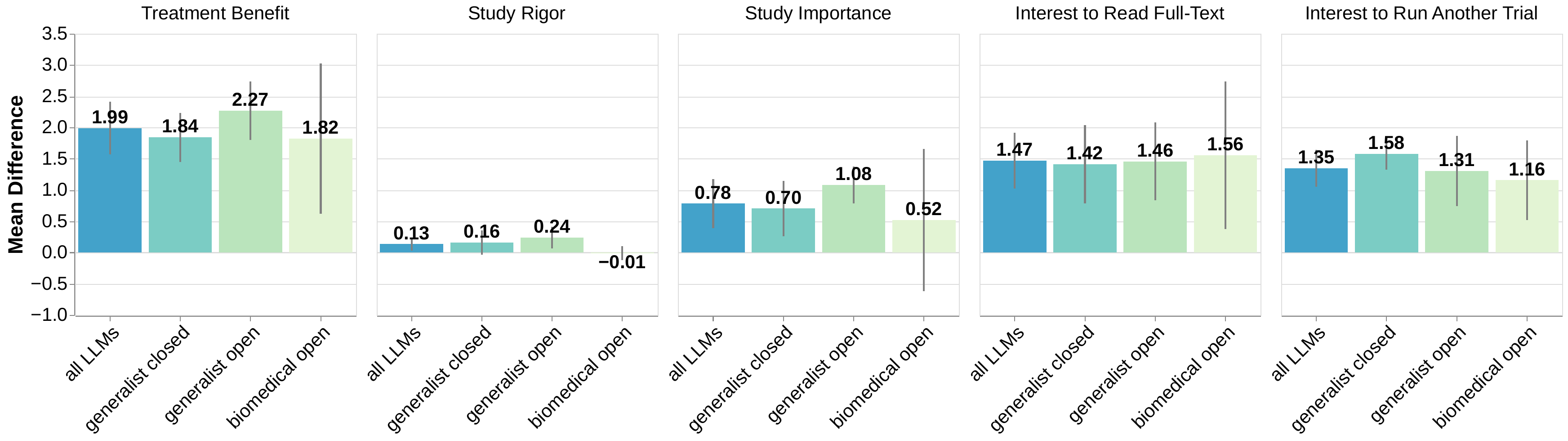}}
\end{figure*}

\begin{figure*}
\floatconts
  {fig:additional_spin_label_regression_benefit}
  {\caption{Coefficients from linear regression models with 95\% CI for each LLM showing how much different LLMs overestimate the treatment effects (benefit of treatment), when abstracts contain `spin' in our semi-synthetic data. In comparison with the reference line, all LLMs except for one model were more susceptible to spin. Med42 70B and Olmo2 Instruct 7B were the most susceptible to spin than others.}}
  {\includegraphics[width=0.8\linewidth]{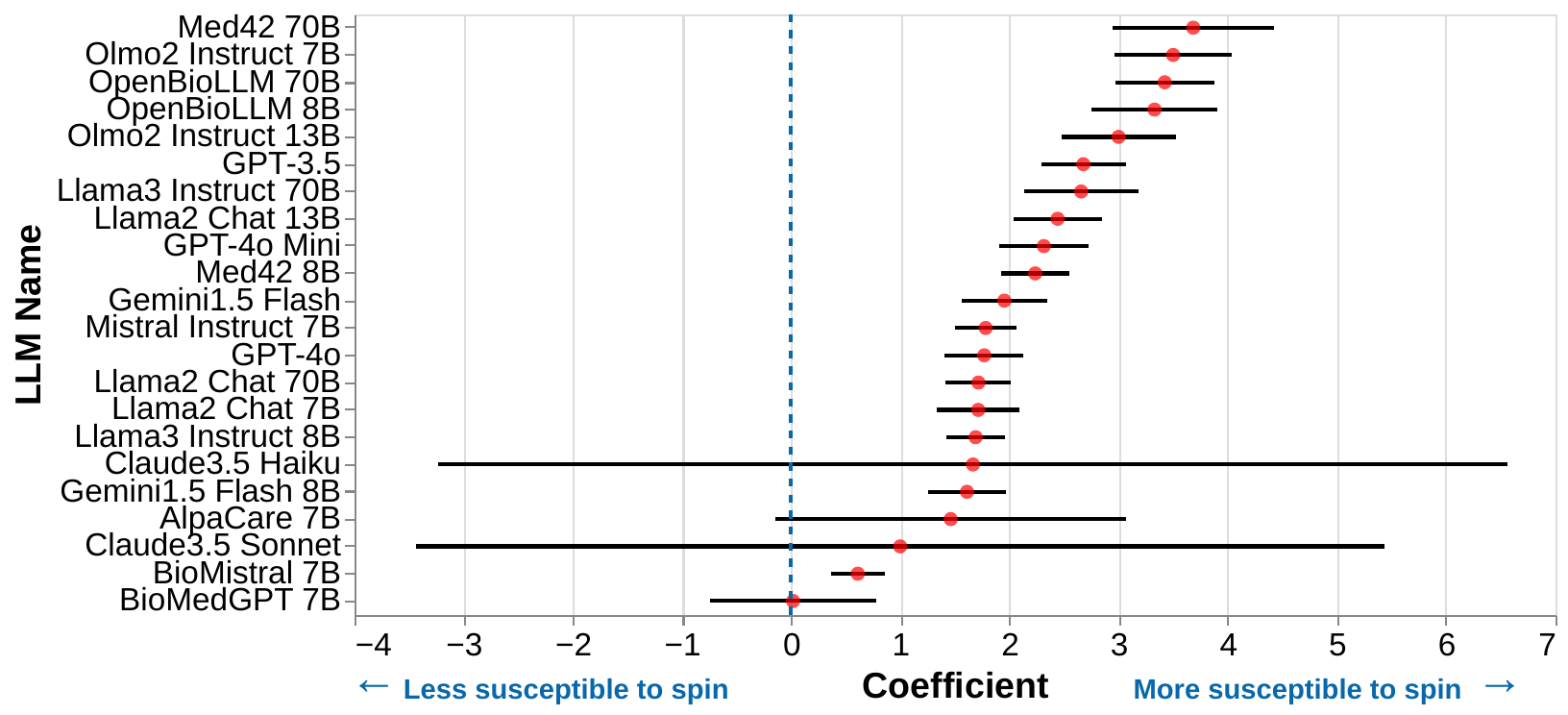}}
\end{figure*}

\paragraph{Simplifying spun abstracts}

Using all 22 models, we generated simplified (plain language) summaries of abstracts from the semi-synthetic dataset. We then used GPT-4o Mini, Claude 3.5 Sonnet, and OpenBioLLM 70B (as in \sectionref{para:simplifying_spun_abstracts}) to assess whether spin propagates through the simplification process. We prompted each evaluator model to answer the interpretation questions using the automatically generated \emph{simplified} abstracts.

The results, shown in \figureref{fig:additional_pls_evaluation}, indicate that spin from the original abstracts persisted in the plain language summaries, potentially influencing LLM interpretation of trial results. Score differences between spun and unspun versions were particularly pronounced for assessments of treatment benefit, interest in full-text review, and motivation to conduct additional trials. Notably, mean difference scores varied more by evaluator model which was similar to what was found in the original evaluation.

For treatment benefit specifically, mean differences were 0.80 for Claude 3.5 Sonnet, 1.69 for GPT-4o Mini, and 1.70 for OpenBioLLM 70B. Mean difference scores for study rigour and importance were on average lower than for the other questions, consistent with findings from the original evaluation. Overall, however, mean difference scores were considerably smaller across questions and evaluator models than those observed in the original evaluation, likely reflecting the model-generated nature of the underlying abstracts.

\begin{figure}
\floatconts
  {fig:additional_pls_evaluation}
  {\caption{Average mean differences of scores from Claude 3.5 Sonnet, GPT-4o Mini, and OpenBioLLM 70B interpreting simplified versions of semi-synthetic abstracts with and without spin generated by 22 LLMs. The error bars indicate 95\% confidence intervals. This plot shows that simplified spun abstracts generated by LLMs also tend to exhibit spin.}}
  {\includegraphics[width=\linewidth]{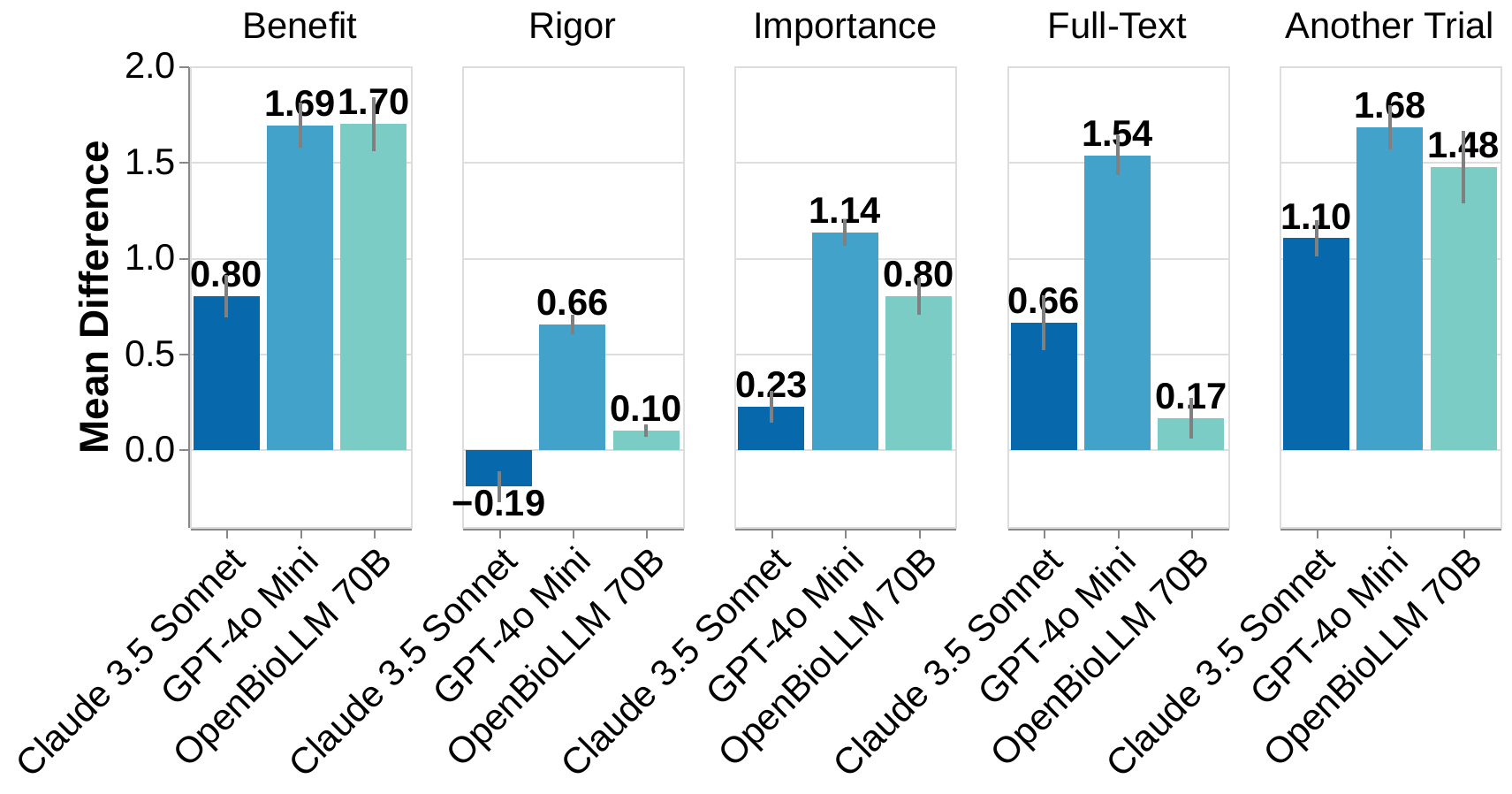}}
\end{figure}

\paragraph{Reducing the effect of spin}

\figureref{fig:additional_interpretation_mitigation_comparisons} shows a comparison of mitigation strategies against baseline prompting for interpreting trial results from the semi-synthetic data. Across all questions, employing mitigation strategies generally reduced mean differences relative to baseline.
Providing reference spin labels produced the largest reduction. For the treatment benefit question, baseline prompting yielded an average mean difference of 1.99 across all LLMs, which decreased to 1.05 when reference labels were included. In contrast, providing \emph{inferred} spin labels offered no improvement for the benefit question, with a mean difference of 2.00. However, this strategy seemed to work with varying levels for the other interpretation questions. Prompting the LLM to first detect spin before answering the interpretation question did mitigate the effect of spin to some extent, though it did not match the reduction achieved by the experiments with the original evaluation set.

\begin{figure*}[htbp]
\floatconts
  {fig:additional_interpretation_mitigation_comparisons}
  {\caption{Average mean differences of scores across all LLMs using different prompting strategies for 5 interpretation questions from our semi-synthetic data. The error bars indicate 95\% confidence intervals.}}
{\includegraphics[width=\linewidth]{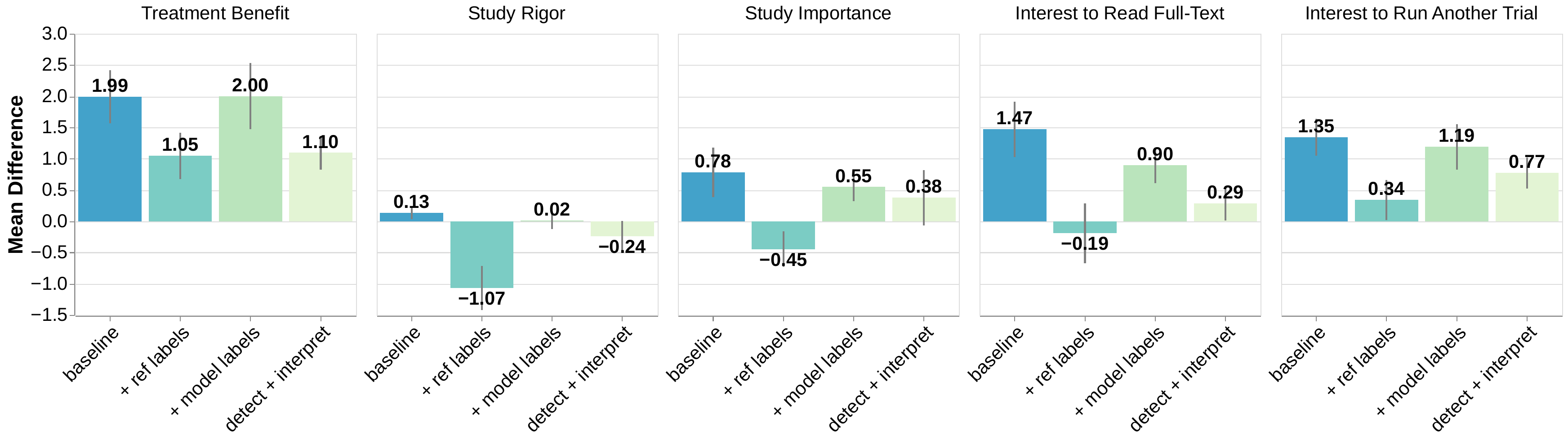}}
\end{figure*}

\section{Related Work}
\label{section:related_work}

\paragraph{Spin in medical literature} Classification and prevalence of spin in medical literature has been studied extensively. 
In this literature, spin is commonly defined in one or more of the following ways: (1) Distorting the interpretation of results to present them more favorably, resulting in misleading conclusions; (2) Discrepancies between results and their favorable interpretation; (3) Attributing causality when the study design does not support it; and (4) Inappropriate extrapolation of results \citep{chiu2017spin}.

A systematic review revealed that spin is more prevalent in trials with nonsignificant primary outcomes and those with a higher risk of bias, such as nonrandomized studies \citep{chiu2017spin}. For instance, $\sim$60\% of RCTs with nonsignificant primary outcomes that were evaluated featured spin in their abstracts. 
Beyond RCTs, spin has also been identified as a common issue in systematic reviews and meta-analyses \citep{yavchitz2016new, nowlin2022spin, qureshi2024development}. Research has demonstrated the presence of spin across various medical disciplines, including oncology \citep{boutron2010reporting, wayant2019evaluation}, psychiatry and psychology \citep{jellison2020evaluation}, dental care \citep{su2023assessment}, wound care \citep{lockyer2013spin}, cardiovascular medicine \citep{khan2019level}, rheumatology \citep{mathieu2012misleading}, obesity research \citep{austin2019evaluation}, and emergency medicine \citep{reynolds2020evaluation}.

\paragraph{NLP for detecting spin} One of the earliest efforts in automated spin detection was by \citet{koroleva2017contribution}, who proposed a pipeline that used rule-based and deep learning methods to identify key entities and comparative sentences (e.g., ``Treatment A was better than treatment B in terms of efficacy.'') to analyze RCT abstracts from PubMed.
This laid the foundation for subsequent research, including the creation of a corpus of biomedical articles annotated for spin \citep{koroleva2018annotating} and the development of DeSpin, a prototype tool designed to assist authors and reviewers in detecting spin in RCT abstracts \citep{koroleva2020despin}. 
DeSpin adopted a combination of rule-based methods and machine learning models---BERT \citep{devlin2018bert}, BioBERT \citep{lee2020biobert}, and SciBERT \citep{beltagy2019scibert}---to extract trial design information and identify spin. 
Our work extends this line of inquiry, evaluating modern LLMs for spin-related tasks.

\paragraph{NLP for simplifying medical texts} Several studies have explored the use of NLP technologies to simplify medical texts, which are often laden with jargon, to make them more accessible to lay readers \citep{ondov2022survey}. Recent advancements in deep learning and LLMs have enabled the simplification of medical abstracts \citep{devaraj2021paragraph, shaib-etal-2023-summarizing} and patient notes, including radiology reports \citep{bala2020patient, jeblick2024chatgpt}. An interactive tool has also been developed to provide simplified summaries of medical articles, assisting lay readers in understanding complex medical texts \citep{august2023paper}. Additionally, efforts have been made to improve evaluation methods for medical text simplification by creating high-quality corpora \citep{devaraj2022evaluating, joseph2023multilingual}. However, existing work does not address spin in medical literature.

\section{Discussion and Conclusions}

We empirically investigated the susceptibility of LLMs to spin in medical articles. 
Specifically, we evaluated 22 LLMs of various sizes and types (generalist and biomedical; closed and open weight) on tasks intended to provide insights regarding the capabilities and limitations of LLMs with respect to handling spin.

In particular, we first assessed the ability of LLMs to detect the presence of spin in abstracts of RCT reports.
LLMs demonstrated moderate to strong performance on this task (average accuracy across LLMs: 0.67). However, few LLMs had accuracy scores of 50\% due to labeling all abstracts as ``spin'' or ``no spin.'' This showed that some LLMs found spin detection in a zero-shot setting challenging and may not be optimally instruction-tuned to follow our intended request. 
Despite being trained on medical literature, biomedical LLMs did not consistently outperform generalist models. Our findings align with recent work showing that medical LLMs fail to consistently improve over their base models in both zero- and few-shot prompting for medical question-answering (QA) tasks \citep{jeong2024medical}.

Next we examined the extent to which LLMs are influenced by spin when tasked with interpreting trial results. 
By comparing numerical interpretation scores generated by LLMs for spun and unspun abstracts, we found that LLMs are generally \emph{more} susceptible to spin than are human experts. 
Across all model types and sizes considered, LLMs interpreted trial results more favorably when spin was present regardless of their performance in detecting spin. 
This susceptibility reveals a significant limitation of LLMs for unbiased interpretation of trial reports. Recent work has touted LLM's mathematical reasoning capabilities \citep{imani2023mathprompter, romera2024mathematical}, but the ability to recognize spin as a discrepancy between numerical results and language demands more in discourse reasoning. 

Finally, we explored whether LLMs might propagate spin, specifically when generating simplified versions of medical abstracts. 
We found that this is indeed a risk: LLMs seem to consistently propagate spin, insofar as simplified versions of spun abstracts were interpreted more favorably by LLM evaluators. 
However, it remains unclear whether LLMs amplify spin beyond the level present in the original text. 

While these findings together suggest that LLMs are indeed susceptible to spin despite strong spin detection performance, we have also proposed and evaluated potential strategies to reduce its impact on LLM outputs. 
Specifically, we found that LLM susceptibility can be reduced via intentional prompting, e.g., providing additional context about the presence of spin, or instructing the model to simultaneously detect and interpret spin, significantly reducing the exaggeration of treatment benefits. The significant benefits of using this joint prompting approach align with existing work on prompting strategies to improve LLM reasoning \citep{qiao2023reasoning}.
Our results suggest that careful prompt engineering may be crucial for developing and designing LLMs to be more reliable for biomedical applications.

Further experiments on larger, lower-quality datasets further proved LLMs' susceptibility to spin across a broader range of abstracts. The demonstrated susceptibility of LLMs to spin warrants careful consideration as these models are increasingly used to analyze and synthesize medical literature. Although LLMs show baseline vulnerability to spin, our findings suggest they can be effectively prompted to detect and mitigate its influence. Future research should focus on developing robust strategies to reduce spin susceptibility, ultimately supporting more accurate evidence synthesis for clinical decision-making.

\subsection*{Limitations}

Our main evaluation relies on a small (but high-quality) dataset of manually selected and edited oncology trials. This may limit the generalizability of our findings. However, we created a larger dataset (but lower-quality) and showed generalizability to trials outside of oncology but did not compare with any human expert baseline.
Future work would ideally explore much larger, more diverse evaluation datasets to deepen our analysis of LLM susceptibility to spin in medical literature and other domains.

Another limitation concerns our assumptions regarding supervision. We primarily evaluated the zero-shot capabilities of LLMs in detecting spin in, interpreting trial results from, and generating simplified versions of published trial results. 
Few-shot prompting or fine-tuning would probably improve performance on these tasks, though would require collecting annotations.

For evaluating simplified versions of abstracts generated by LLMs, we used two proprietary, closed-source LLMs within an LLM-as-evaluator framework. This study does not compare LLM-generated evaluation scores with those from lay human evaluators, limiting our ability to confirm alignment between automated assessments and human judgments. Future work should incorporate human evaluations to further validate these findings.



\acks{This research was supported by the National Institutes of Health (NIH) grant 1R01LM014600-01 and National Science Foundation (NSF) grant IIS-2145479.}

\bibliography{bibliography}

@article{hopewell2008consort,
  title={CONSORT for reporting randomized controlled trials in journal and conference abstracts: explanation and elaboration},
  author={Hopewell, Sally and Clarke, Mike and Moher, David and Wager, Elizabeth and Middleton, Philippa and Altman, Douglas G and Schulz, Kenneth F and Consort Group},
  journal={PLoS medicine},
  volume={5},
  number={1},
  pages={e20},
  year={2008},
  publisher={Public Library of Science San Francisco, USA}
}

@article{day2020open,
  title={Open to the public: paywalls and the public rationale for open access medical research publishing},
  author={Day, Suzanne and Rennie, Stuart and Luo, Danyang and Tucker, Joseph D},
  journal={Research involvement and engagement},
  volume={6},
  pages={1--7},
  year={2020},
  publisher={Springer}
}

@article{thirunavukarasu2023large,
  title={Large language models in medicine},
  author={Thirunavukarasu, Arun James and Ting, Darren Shu Jeng and Elangovan, Kabilan and Gutierrez, Laura and Tan, Ting Fang and Ting, Daniel Shu Wei},
  journal={Nature medicine},
  volume={29},
  number={8},
  pages={1930--1940},
  year={2023},
  publisher={Nature Publishing Group US New York}
}

@inproceedings{shaib-etal-2023-summarizing,
    title = "Summarizing, Simplifying, and Synthesizing Medical Evidence using {GPT}-3 (with Varying Success)",
    author = "Shaib, Chantal  and
      Li, Millicent  and
      Joseph, Sebastian  and
      Marshall, Iain  and
      Li, Junyi Jessy  and
      Wallace, Byron",
    editor = "Rogers, Anna  and
      Boyd-Graber, Jordan  and
      Okazaki, Naoaki",
    booktitle = "Proceedings of the Annual Meeting of the Association for Computational Linguistics (ACL)",
    month = jul,
    year = "2023",
    address = "Toronto, Canada",
    publisher = "Association for Computational Linguistics",
    url = "https://aclanthology.org/2023.acl-short.119/",
    doi = "10.18653/v1/2023.acl-short.119",
    pages = "1387--1407",
}

@article{piwowar2018state,
  title={The state of OA: a large-scale analysis of the prevalence and impact of Open Access articles},
  author={Piwowar, Heather and Priem, Jason and Larivi{\`e}re, Vincent and Alperin, Juan Pablo and Matthias, Lisa and Norlander, Bree and Farley, Ashley and West, Jevin and Haustein, Stefanie},
  journal={PeerJ},
  volume={6},
  pages={e4375},
  year={2018},
  publisher={PeerJ Inc.}
}

@inproceedings{devaraj2022evaluating,
  title={Evaluating factuality in text simplification},
  author={Devaraj, Ashwin and Sheffield, William and Wallace, Byron C and Li, Junyi Jessy},
  booktitle={Proceedings of the conference of the Association for Computational Linguistics (ACL)},
  volume={2022},
  pages={7331},
  year={2022},
}

@article{smith2017knowledge,
  title={Knowledge sharing in global health research--the impact, uptake and cost of open access to scholarly literature},
  author={Smith, Elise and Haustein, Stefanie and Mongeon, Philippe and Shu, Fei and Ridde, Val{\'e}ry and Larivi{\`e}re, Vincent},
  journal={Health Research Policy and Systems},
  volume={15},
  pages={1--10},
  year={2017},
  publisher={Springer}
}

@article{barry2001family,
  title={Family physicians' use of medical abstracts to guide decision making: style or substance?},
  author={Barry, Henry C and Ebell, Mark H and Shaughnessy, Allen F and Slawson, David C and Nietzke, Fern},
  journal={The Journal of the American Board of Family Practice},
  volume={14},
  number={6},
  pages={437--442},
  year={2001},
  publisher={Am Board Family Med}
}

@article{islamaj2009understanding,
  title={Understanding PubMed{\textregistered} user search behavior through log analysis},
  author={Islamaj Dogan, Rezarta and Murray, G Craig and N{\'e}v{\'e}ol, Aur{\'e}lie and Lu, Zhiyong},
  journal={Database},
  volume={2009},
  pages={bap018},
  year={2009},
  publisher={Oxford University Press}
}

@article{boutron2010reporting,
  title={Reporting and interpretation of randomized controlled trials with statistically nonsignificant results for primary outcomes},
  author={Boutron, Isabelle and Dutton, Susan and Ravaud, Philippe and Altman, Douglas G},
  journal={Jama},
  volume={303},
  number={20},
  pages={2058--2064},
  year={2010},
  publisher={American Medical Association}
}

@article{chiu2017spin,
  title={‘Spin’in published biomedical literature: a methodological systematic review},
  author={Chiu, Kellia and Grundy, Quinn and Bero, Lisa},
  journal={PLoS Biology},
  volume={15},
  number={9},
  pages={e2002173},
  year={2017},
  publisher={Public Library of Science San Francisco, CA USA}
}

@article{jellison2020evaluation,
  title={Evaluation of spin in abstracts of papers in psychiatry and psychology journals},
  author={Jellison, Samuel and Roberts, Will and Bowers, Aaron and Combs, Tyler and Beaman, Jason and Wayant, Cole and Vassar, Matt},
  journal={BMJ evidence-based medicine},
  volume={25},
  number={5},
  pages={178--181},
  year={2020},
  publisher={Royal Society of Medicine}
}

@article{nowlin2022spin,
  title={Spin in Abstracts of Systematic Reviews and Meta-analyses of Melanoma Therapies: Cross-sectional Analysis},
  author={Nowlin, Ross and Wirtz, Alexis and Wenger, David and Ottwell, Ryan and Cook, Courtney and Arthur, Wade and Sallee, Brigitte and Levin, Jarad and Hartwell, Micah and Wright, Drew and others},
  journal={JMIR dermatology},
  volume={5},
  number={1},
  pages={e33996},
  year={2022},
  publisher={JMIR Publications Toronto, Canada}
}

@article{wayant2019evaluation,
  title={Evaluation of spin in oncology clinical trials},
  author={Wayant, C and Margalski, D and Vaughn, K and Vassar, M},
  journal={Critical Reviews in Oncology/Hematology},
  volume={144},
  pages={102821},
  year={2019},
  publisher={Elsevier}
}

@article{sumner2016exaggerations,
  title={Exaggerations and caveats in press releases and health-related science news},
  author={Sumner, Petroc and Vivian-Griffiths, Solveiga and Boivin, Jacky and Williams, Andrew and Bott, Lewis and Adams, Rachel and Venetis, Christos A and Whelan, Leanne and Hughes, Bethan and Chambers, Christopher D},
  journal={PloS one},
  volume={11},
  number={12},
  pages={e0168217},
  year={2016},
  publisher={Public Library of Science San Francisco, CA USA}
}

@book{prasad2020malignant,
  title={Malignant: how bad policy and bad evidence harm people with Cancer},
  author={Prasad, Vinayak K},
  year={2020},
  publisher={JHU Press}
}

@article{yavchitz2016new,
  title={A new classification of spin in systematic reviews and meta-analyses was developed and ranked according to the severity},
  author={Yavchitz, Am{\'e}lie and Ravaud, Philippe and Altman, Douglas G and Moher, David and Hrobjartsson, Asbj{\o}rn and Lasserson, Toby and Boutron, Isabelle},
  journal={Journal of clinical epidemiology},
  volume={75},
  pages={56--65},
  year={2016},
  publisher={Elsevier}
}

@article{boutron2014impact,
  title={Impact of spin in the abstracts of articles reporting results of randomized controlled trials in the field of cancer: the SPIIN randomized controlled trial},
  author={Boutron, Isabelle and Altman, Douglas G and Hopewell, Sally and Vera-Badillo, Francisco and Tannock, Ian and Ravaud, Philippe},
  journal={Journal of Clinical Oncology},
  volume={32},
  number={36},
  pages={4120--4126},
  year={2014},
  publisher={American Society of Clinical Oncology}
}

@article{team2024gemini,
  title={Gemini 1.5: Unlocking multimodal understanding across millions of tokens of context},
  author={Team, Gemini and Georgiev, Petko and Lei, Ving Ian and Burnell, Ryan and Bai, Libin and Gulati, Anmol and Tanzer, Garrett and Vincent, Damien and Pan, Zhufeng and Wang, Shibo and others},
  journal={arXiv preprint arXiv:2403.05530},
  year={2024}
}

@article{hurst2024gpt,
  title={Gpt-4o system card},
  author={Hurst, Aaron and Lerer, Adam and Goucher, Adam P and Perelman, Adam and Ramesh, Aditya and Clark, Aidan and Ostrow, AJ and Welihinda, Akila and Hayes, Alan and Radford, Alec and others},
  journal={arXiv preprint arXiv:2410.21276},
  year={2024}
}

@article{openai20244o,
  title={4o mini: Advancing cost-efficient intelligence, 2024},
  author={OpenAI, Gpt},
  journal={URL: https://openai. com/index/gpt-4o-mini-advancing-cost-efficient-intelligence},
  year={2024}
}

@article{touvron2023llama,
  title={Llama 2: Open foundation and fine-tuned chat models},
  author={Touvron, Hugo and Martin, Louis and Stone, Kevin and Albert, Peter and Almahairi, Amjad and Babaei, Yasmine and Bashlykov, Nikolay and Batra, Soumya and Bhargava, Prajjwal and Bhosale, Shruti and others},
  journal={arXiv preprint arXiv:2307.09288},
  year={2023}
}

@article{meta2024introducing,
  title={Introducing meta llama 3: The most capable openly available llm to date},
  author={Meta, AI},
  journal={Meta AI},
  year={2024}
}

@article{jiang2023mistral,
  title={Mistral 7B},
  author={Jiang, Albert Q and Sablayrolles, Alexandre and Mensch, Arthur and Bamford, Chris and Chaplot, Devendra Singh and Casas, Diego de las and Bressand, Florian and Lengyel, Gianna and Lample, Guillaume and Saulnier, Lucile and others},
  journal={arXiv preprint arXiv:2310.06825},
  year={2023}
}

@article{olmo20242,
  title={2 OLMo 2 Furious},
  author={OLMo, Team and Walsh, Pete and Soldaini, Luca and Groeneveld, Dirk and Lo, Kyle and Arora, Shane and Bhagia, Akshita and Gu, Yuling and Huang, Shengyi and Jordan, Matt and others},
  journal={arXiv preprint arXiv:2501.00656},
  year={2024}
}

@article{zhang2023alpacare,
  title={Alpacare: Instruction-tuned large language models for medical application},
  author={Zhang, Xinlu and Tian, Chenxin and Yang, Xianjun and Chen, Lichang and Li, Zekun and Petzold, Linda Ruth},
  journal={arXiv preprint arXiv:2310.14558},
  year={2023}
}

@article{luo2023biomedgpt,
  title={Biomedgpt: Open multimodal generative pre-trained transformer for biomedicine},
  author={Luo, Yizhen and Zhang, Jiahuan and Fan, Siqi and Yang, Kai and Wu, Yushuai and Qiao, Mu and Nie, Zaiqing},
  journal={arXiv preprint arXiv:2308.09442},
  year={2023}
}

@article{labrak2024biomistral,
  title={Biomistral: A collection of open-source pretrained large language models for medical domains},
  author={Labrak, Yanis and Bazoge, Adrien and Morin, Emmanuel and Gourraud, Pierre-Antoine and Rouvier, Mickael and Dufour, Richard},
  journal={arXiv preprint arXiv:2402.10373},
  year={2024}
}

@article{christophe2024med42,
  title={Med42-v2: A suite of clinical llms},
  author={Christophe, Cl{\'e}ment and Kanithi, Praveen K and Raha, Tathagata and Khan, Shadab and Pimentel, Marco AF},
  journal={arXiv preprint arXiv:2408.06142},
  year={2024}
}

@misc{pal2024openbiollms,
  title={Openbiollms: Advancing open-source large language models for healthcare and life sciences},
  author={Pal, Malaikannan Sankarasubbu Ankit and Sankarasubbu, Malaikannan},
  year={2024}
}

@manual{openai2025gpt35turbo,
  title        = {Models},
  author       = {OpenAI},
  year         = {2025},
  url          = {https://platform.openai.com/docs/models/gpt-3-5-turbo},
  note         = {Accessed: 2025-01-17}
}

@inproceedings{joseph2024factpico,
    title = "{F}act{PICO}: Factuality Evaluation for Plain Language Summarization of Medical Evidence",
    author = {Joseph, Sebastian  and
      Chen, Lily  and
      Trienes, Jan  and
      G{\"o}ke, Hannah  and
      Coers, Monika  and
      Xu, Wei  and
      Wallace, Byron  and
      Li, Junyi Jessy},
    editor = "Ku, Lun-Wei  and
      Martins, Andre  and
      Srikumar, Vivek",
    booktitle = "Proceedings of the 62nd Annual Meeting of the Association for Computational Linguistics (Volume 1: Long Papers)",
    month = aug,
    year = "2024",
    address = "Bangkok, Thailand",
    publisher = "Association for Computational Linguistics",
    url = "https://aclanthology.org/2024.acl-long.459/",
    doi = "10.18653/v1/2024.acl-long.459",
    pages = "8437--8464"
}

@article{easterbrook1991publication,
  title={Publication bias in clinical research},
  author={Easterbrook, Phillipa J and Gopalan, Ramana and Berlin, JA and Matthews, David R},
  journal={The Lancet},
  volume={337},
  number={8746},
  pages={867--872},
  year={1991},
  publisher={Elsevier}
}

@article{begg1988publication,
  title={Publication bias: a problem in interpreting medical data},
  author={Begg, Colin B and Berlin, Jesse A},
  journal={Journal of the Royal Statistical Society Series A: Statistics in Society},
  volume={151},
  number={3},
  pages={419--445},
  year={1988},
  publisher={Oxford University Press}
}

@article{devito2019catalogue,
  title={Catalogue of bias: publication bias},
  author={DeVito, Nicholas J and Goldacre, Ben},
  journal={BMJ Evidence-Based Medicine},
  volume={24},
  number={2},
  pages={53--54},
  year={2019},
  publisher={Royal Society of Medicine}
}

@ARTICLE{dickersin1993publication,
  title     = "Publication bias: the problem that won't go away",
  author    = "Dickersin, K and Min, Y I",
  journal   = "Ann. N. Y. Acad. Sci.",
  publisher = "Wiley",
  volume    =  703,
  number    =  1,
  pages     = "135--46; discussion 146--8",
  month     =  dec,
  year      =  1993,
  copyright = "http://onlinelibrary.wiley.com/termsAndConditions\#vor",
}

@article{dickersin1990existence,
  title={The existence of publication bias and risk factors for its occurrence},
  author={Dickersin, Kay},
  journal={Jama},
  volume={263},
  number={10},
  pages={1385--1389},
  year={1990},
  publisher={American Medical Association}
}

@article{jeblick2024chatgpt,
  title={ChatGPT makes medicine easy to swallow: an exploratory case study on simplified radiology reports},
  author={Jeblick, Katharina and Schachtner, Balthasar and Dexl, Jakob and Mittermeier, Andreas and St{\"u}ber, Anna Theresa and Topalis, Johanna and Weber, Tobias and Wesp, Philipp and Sabel, Bastian Oliver and Ricke, Jens and others},
  journal={European radiology},
  volume={34},
  number={5},
  pages={2817--2825},
  year={2024},
  publisher={Springer}
}

@article{picton2025assessing,
  title={Assessing AI simplification of medical texts: readability and content fidelity},
  author={Picton, Bryce and Andalib, Saman and Spina, Aidin and Camp, Brandon and Solomon, Sean S and Liang, Jason and Chen, Patrick M and Chen, Jefferson W and Hsu, Frank P and Oh, Michael Y},
  journal={International Journal of Medical Informatics},
  volume={195},
  pages={105743},
  year={2025},
  publisher={Elsevier}
}

@article{august2023paper,
  title={Paper plain: Making medical research papers approachable to healthcare consumers with natural language processing},
  author={August, Tal and Wang, Lucy Lu and Bragg, Jonathan and Hearst, Marti A and Head, Andrew and Lo, Kyle},
  journal={ACM Transactions on Computer-Human Interaction},
  volume={30},
  number={5},
  pages={1--38},
  year={2023},
  publisher={ACM New York, NY}
}

@article{osborne2002double,
  title={Double-blind, randomized trial comparing the efficacy and tolerability of fulvestrant versus anastrozole in postmenopausal women with advanced breast cancer progressing on prior endocrine therapy: results of a North American trial},
  author={Osborne, CK and Pippen, J and Jones, SE and Parker, LM and Ellis, M and Come, S and Gertler, SZ and May, JT and Burton, G and Dimery, I and others},
  journal={Journal of Clinical Oncology},
  volume={20},
  number={16},
  pages={3386--3395},
  year={2002},
  publisher={American Society of Clinical Oncology}
}

@article{marchand2017readability,
  title={What is readability and why should content editors care about it},
  author={Marchand, Lisa},
  journal={Center for Plain Language. https://centerforplainlanguage. org/what-isreadability},
  year={2017}
}

@article{kutner2006health,
  title={The Health Literacy of America's Adults: Results from the 2003 National Assessment of Adult Literacy. NCES 2006-483.},
  author={Kutner, Mark and Greenburg, Elizabeth and Jin, Ying and Paulsen, Christine},
  journal={National Center for education statistics},
  year={2006},
  publisher={ERIC}
}

@book{weiss2007health,
  title={Health literacy and patient safety: Help patients understand. Manual for clinicians},
  author={Weiss, Barry D},
  year={2007},
  publisher={American Medical Association Foundation}
}

@article{us2009simply,
  title={Simply Put: A guide for creating easy-to-understand materials},
  author={US Department of Health and Human Services and others},
  year={2009}
}

@misc{brega2015ahrq,
  title={AHRQ health literacy universal precautions toolkit},
  author={Brega, AGBJ and Barnard, J and Mabachi, NM and Weiss, B and DeWalt, D and Brach, C and Cifuentes, M and Albright, K and West, D},
  year={2015}
}

@article{goldsack2025leveraging,
  title={Leveraging Large Language Models for Zero-shot Lay Summarisation in Biomedicine and Beyond},
  author={Goldsack, Tomas and Scarton, Carolina and Lin, Chenghua},
  journal={arXiv preprint arXiv:2501.05224},
  year={2025}
}

@article{boutron2019three,
  title={Three randomized controlled trials evaluating the impact of “spin” in health news stories reporting studies of pharmacologic treatments on patients’/caregivers’ interpretation of treatment benefit},
  author={Boutron, Isabelle and Haneef, Romana and Yavchitz, Am{\'e}lie and Baron, Gabriel and Novack, John and Oransky, Ivan and Schwitzer, Gary and Ravaud, Philippe},
  journal={BMC medicine},
  volume={17},
  pages={1--10},
  year={2019},
  publisher={Springer}
}

@article{wei2022chain,
  title={Chain-of-thought prompting elicits reasoning in large language models},
  author={Wei, Jason and Wang, Xuezhi and Schuurmans, Dale and Bosma, Maarten and Xia, Fei and Chi, Ed and Le, Quoc V and Zhou, Denny and others},
  journal={Advances in neural information processing systems},
  volume={35},
  pages={24824--24837},
  year={2022}
}

@inproceedings{koroleva2017contribution,
  title={On the Contribution of Specific Entity Detection in Comparative Constructions to Automatic Spin Detection in Biomedical Scientific Publications},
  author={Koroleva, Anna and Paroubek, Patrick},
  booktitle={Language and Technology Conference},
  pages={304--317},
  year={2017},
  organization={Springer}
}

@inproceedings{koroleva2020despin,
  title={DeSpin: a prototype system for detecting spin in biomedical publications},
  author={Koroleva, Anna and Kamath, Sanjay and Bossuyt, Patrick MM and Paroubek, Patrick},
  booktitle={roceedings of the BioNLP 2020 workshop},
  pages={49--59},
  year={2020},
  organization={Association for Computational Linguistics}
}

@inproceedings{koroleva2018annotating,
  title={Annotating Spin in Biomedical Scientific Publications: the case of Random Controlled Trials (RCTs)},
  author={Koroleva, Anna and Paroubek, Patrick},
  booktitle={Proceedings of the Eleventh International Conference on Language Resources and Evaluation (LREC 2018)},
  year={2018}
}

@article{devlin2018bert,
  title={Bert: Pre-training of deep bidirectional transformers for language understanding},
  author={Devlin, Jacob},
  journal={arXiv preprint arXiv:1810.04805},
  year={2018}
}

@article{lee2020biobert,
  title={BioBERT: a pre-trained biomedical language representation model for biomedical text mining},
  author={Lee, Jinhyuk and Yoon, Wonjin and Kim, Sungdong and Kim, Donghyeon and Kim, Sunkyu and So, Chan Ho and Kang, Jaewoo},
  journal={Bioinformatics},
  volume={36},
  number={4},
  pages={1234--1240},
  year={2020},
  publisher={Oxford University Press}
}

@inproceedings{beltagy2019scibert,
    title = "{S}ci{BERT}: A Pretrained Language Model for Scientific Text",
    author = "Beltagy, Iz  and
      Lo, Kyle  and
      Cohan, Arman",
    editor = "Inui, Kentaro  and
      Jiang, Jing  and
      Ng, Vincent  and
      Wan, Xiaojun",
    booktitle = "Proceedings of the 2019 Conference on Empirical Methods in Natural Language Processing and the 9th International Joint Conference on Natural Language Processing (EMNLP-IJCNLP)",
    month = nov,
    year = "2019",
    address = "Hong Kong, China",
    publisher = "Association for Computational Linguistics",
    url = "https://aclanthology.org/D19-1371/",
    doi = "10.18653/v1/D19-1371",
    pages = "3615--3620"
}

@inproceedings{devaraj2021paragraph,
  title={Paragraph-level simplification of medical texts},
  author={Devaraj, Ashwin and Wallace, Byron C and Marshall, Iain J and Li, Junyi Jessy},
  booktitle={Proceedings of the conference. Association for Computational Linguistics. North American Chapter. Meeting},
  volume={2021},
  pages={4972},
  year={2021},
  organization={NIH Public Access}
}

@article{ondov2022survey,
  title={A survey of automated methods for biomedical text simplification},
  author={Ondov, Brian and Attal, Kush and Demner-Fushman, Dina},
  journal={Journal of the American Medical Informatics Association},
  volume={29},
  number={11},
  pages={1976--1988},
  year={2022},
  publisher={Oxford University Press}
}

@article{joseph2023multilingual,
  title={Multilingual simplification of medical texts},
  author={Joseph, Sebastian and Kazanas, Kathryn and Reina, Keziah and Ramanathan, Vishnesh J and Xu, Wei and Wallace, Byron C and Li, Junyi Jessy},
  journal={arXiv preprint arXiv:2305.12532},
  year={2023}
}

@article{su2023assessment,
  title={Assessment of spin in the abstracts of randomized controlled trials in dental caries with statistically nonsignificant results for primary outcomes: A methodological study},
  author={Su, Naichuan and Van Der Linden, Michiel W and Faggion Jr, Clovis M and Van Der Heijden, Geert JMG},
  journal={Caries Research},
  volume={57},
  number={5-6},
  pages={553--562},
  year={2023},
  publisher={S. Karger AG}
}

@article{lockyer2013spin,
  title={"Spin" in wound care research: the reporting and interpretation of randomized controlled trials with statistically non-significant primary outcome results or unspecified primary outcomes},
  author={Lockyer, Suzanne and Hodgson, Rob and Dumville, Jo C and Cullum, Nicky},
  journal={Trials},
  volume={14},
  pages={1--10},
  year={2013},
  publisher={Springer}
}

@article{khan2019level,
  title={Level and prevalence of spin in published cardiovascular randomized clinical trial reports with statistically nonsignificant primary outcomes: a systematic review},
  author={Khan, Muhammad Shahzeb and Lateef, Noman and Siddiqi, Tariq Jamal and Rehman, Karim Abdur and Alnaimat, Saed and Khan, Safi U and Riaz, Haris and Murad, M Hassan and Mandrola, John and Doukky, Rami and others},
  journal={JAMA network open},
  volume={2},
  number={5},
  pages={e192622--e192622},
  year={2019},
  publisher={American Medical Association}
}

@article{mathieu2012misleading,
  title={Misleading abstract conclusions in randomized controlled trials in rheumatology: comparison of the abstract conclusions and the results section},
  author={Mathieu, Sylvain and Giraudeau, Bruno and Soubrier, Martin and Ravaud, Philippe},
  journal={Joint Bone Spine},
  volume={79},
  number={3},
  pages={262--267},
  year={2012},
  publisher={Elsevier}
}

@article{austin2019evaluation,
  title={Evaluation of spin within abstracts in obesity randomized clinical trials: a cross-sectional review},
  author={Austin, Jennifer and Smith, Christopher and Natarajan, Kavita and Som, Mousumi and Wayant, Cole and Vassar, Matt},
  journal={Clinical obesity},
  volume={9},
  number={2},
  pages={e12292},
  year={2019},
  publisher={Wiley Online Library}
}

@article{reynolds2020evaluation,
  title={Evaluation of spin in the abstracts of emergency medicine randomized controlled trials},
  author={Reynolds-Vaughn, Victoria and Riddle, Jonathan and Brown, Jamin and Schiesel, Michael and Wayant, Cole and Vassar, Matt},
  journal={Annals of emergency medicine},
  volume={75},
  number={3},
  pages={423--431},
  year={2020},
  publisher={Elsevier}
}

@article{qureshi2024development,
  title={Development and Evaluation of a Framework for Identifying and Addressing Spin for Harms in Systematic Reviews of Interventions},
  author={Qureshi, Riaz and Naaman, Kevin and Quan, Nicolas G and Mayo-Wilson, Evan and Page, Matthew J and Cornelius, Victoria and Chou, Roger and Boutron, Isabelle and Golder, Su and Bero, Lisa and others},
  journal={Annals of internal medicine},
  volume={177},
  number={8},
  pages={1089--1098},
  year={2024},
  publisher={American College of Physicians}
}

@article{bala2020patient,
  title={Patient perception of plain-language medical notes generated using artificial intelligence software: pilot mixed-methods study},
  author={Bala, Sandeep and Keniston, Angela and Burden, Marisha and others},
  journal={JMIR formative research},
  volume={4},
  number={6},
  pages={e16670},
  year={2020},
  publisher={JMIR Publications Inc., Toronto, Canada}
}

@article{romera2024mathematical,
  title={Mathematical discoveries from program search with large language models},
  author={Romera-Paredes, Bernardino and Barekatain, Mohammadamin and Novikov, Alexander and Balog, Matej and Kumar, M Pawan and Dupont, Emilien and Ruiz, Francisco JR and Ellenberg, Jordan S and Wang, Pengming and Fawzi, Omar and others},
  journal={Nature},
  volume={625},
  number={7995},
  pages={468--475},
  year={2024},
  publisher={Nature Publishing Group UK London}
}

@inproceedings{imani2023mathprompter,
    title = "{M}ath{P}rompter: Mathematical Reasoning using Large Language Models",
    author = "Imani, Shima  and
      Du, Liang  and
      Shrivastava, Harsh",
    editor = "Sitaram, Sunayana  and
      Beigman Klebanov, Beata  and
      Williams, Jason D",
    booktitle = "Proceedings of the 61st Annual Meeting of the Association for Computational Linguistics (Volume 5: Industry Track)",
    month = jul,
    year = "2023",
    address = "Toronto, Canada",
    publisher = "Association for Computational Linguistics",
    url = "https://aclanthology.org/2023.acl-industry.4/",
    doi = "10.18653/v1/2023.acl-industry.4",
    pages = "37--42",
}

@inproceedings{qiao2023reasoning,
    title = "Reasoning with Language Model Prompting: A Survey",
    author = "Qiao, Shuofei  and
      Ou, Yixin  and
      Zhang, Ningyu  and
      Chen, Xiang  and
      Yao, Yunzhi  and
      Deng, Shumin  and
      Tan, Chuanqi  and
      Huang, Fei  and
      Chen, Huajun",
    editor = "Rogers, Anna  and
      Boyd-Graber, Jordan  and
      Okazaki, Naoaki",
    booktitle = "Proceedings of the 61st Annual Meeting of the Association for Computational Linguistics (Volume 1: Long Papers)",
    month = jul,
    year = "2023",
    address = "Toronto, Canada",
    publisher = "Association for Computational Linguistics",
    url = "https://aclanthology.org/2023.acl-long.294/",
    doi = "10.18653/v1/2023.acl-long.294",
    pages = "5368--5393"
}

@article{marcelo2013comparison,
  title={A comparison of the accuracy of clinical decisions based on full-text articles and on journal abstracts alone: a study among residents in a tertiary care hospital},
  author={Marcelo, Alvin and Gavino, Alex and Isip-Tan, Iris Thiele and Apostol-Nicodemus, Leilanie and Mesa-Gaerlan, Faith Joan and Firaza, Paul Nimrod and Faustorilla, John Francis and Callaghan, Fiona M and Fontelo, Paul},
  journal={BMJ Evidence-Based Medicine},
  volume={18},
  number={2},
  pages={48--53},
  year={2013},
  publisher={Royal Society of Medicine}
}

@article{christakis2000physicians,
  title={Do physicians judge a study by its cover?: An investigation of journal attribution bias},
  author={Christakis, Dimitri A and Saint, Sanjay and Saha, Somnath and Elmore, Joann G and Welsh, Deborah E and Baker, Paul and Koepsell, Thomas D},
  journal={Journal of clinical epidemiology},
  volume={53},
  number={8},
  pages={773--778},
  year={2000},
  publisher={Elsevier}
}

@article{berwanger2009quality,
  title={The quality of reporting of trial abstracts is suboptimal: survey of major general medical journals},
  author={Berwanger, Otavio and Ribeiro, Rodrigo A and Finkelsztejn, Alessandro and Watanabe, Marcelo and Suzumura, Erica A and Duncan, Bruce B and Devereaux, Phillip J and Cook, Deborah},
  journal={Journal of clinical epidemiology},
  volume={62},
  number={4},
  pages={387--392},
  year={2009},
  publisher={Elsevier}
}

@inproceedings{jeong2024medical,
    title = "Medical Adaptation of Large Language and Vision-Language Models: Are We Making Progress?",
    author = "Jeong, Daniel P  and
      Garg, Saurabh  and
      Lipton, Zachary Chase  and
      Oberst, Michael",
    editor = "Al-Onaizan, Yaser  and
      Bansal, Mohit  and
      Chen, Yun-Nung",
    booktitle = "Proceedings of the 2024 Conference on Empirical Methods in Natural Language Processing",
    month = nov,
    year = "2024",
    address = "Miami, Florida, USA",
    publisher = "Association for Computational Linguistics",
    url = "https://aclanthology.org/2024.emnlp-main.677/",
    doi = "10.18653/v1/2024.emnlp-main.677",
    pages = "12143--12170"
}

\appendix

\section{LLM Details \& Compute}

We used a total of 4 x NVIDIA A100 GPUs to conduct our experiments.

\paragraph{Generalist closed-source/proprietary:} We used API interfaces provided by OpenAI, Anthropic, and Google to interact with their proprietary language models. For OpenAI models, we used GPT-4o (\verb|gpt-4o-2024-08-06|), GPT-4o-mini (\verb|gpt-4o-mini|), and GPT-3.5 (\verb|gpt-3.5-turbo-0125|). For Anthropic’s Claude models, we tested Claude 3.5 Haiku (\verb|claude-3-5-haiku-20241022|) and Claude 3.5 Sonnet (\verb|claude-3-5-sonnet-20241022|). For Google’s Gemini models, we used Gemini 1.5 Flash (\verb|gemini-1.5-flash|) and Gemini 1.5 Flash 8B (\verb|gemini-1.5-flash-8b|). Across all models and experiments, we set the temperature parameter to 0 to ensure deterministic outputs.

\paragraph{Generalist open-weight:} All the generalist open-weight models were downloaded from HuggingFace's Model Hub\footnote{\url{https://huggingface.co/models}}. We set the parameter for \verb|do_sample| as false for all experiments to ensure deterministic outputs.

\paragraph{Biomedical open-weight:} Similar to the generalist open-weight models, the biomedical models were also downloaded via HuggingFace. The parameter for \verb|do_sample| was also set to false for all experiments.

\section{Zero-Shot Prompts}
\label{appendix:prompts}

\subsection{Detection of Spin}
\label{appendix:detection_prompt}

We provide the exact prompt used in our study for detecting spin below to ensure reproducibility. The prompt was designed to elicit a binary answer of a given abstract.

\begin{tcolorbox}[colback=gray!10, colframe=gray!50, arc=4mm, breakable, enhanced, boxrule=0.4mm, toptitle=0mm, bottomtitle=0mm]
\small
\texttt{Spin or misrepresentation of study findings can be used to influence, positively, the interpretation of statistically nonsignificant randomized controlled trials (RCTs), for example, by emphasizing the apparent benefit of a secondary outcome or findings from a subgroup of patients.}\\
\texttt{Does the following clinical trial abstract contain spin (yes/no)?}\\
\texttt{Answer only with `yes' or `no'. Do not provide any explanations.}\\

\texttt{Abstract: \textbf{\{ABSTRACT\}}}
\end{tcolorbox}

\subsection{Interpretation of Trial Results} 
\label{appendix:interpretation_prompts}

Below, we provide the exact prompt used to guide the LLM in interpreting the results of a given abstract.

\begin{tcolorbox}[colback=gray!10, colframe=gray!50, arc=4mm, breakable, enhanced, boxrule=0.4mm, toptitle=0mm, bottomtitle=0mm]
\small
\texttt{This abstract describes a study evaluating a treatment in the field of cancer published in a peer-reviewed journal.}\\
\texttt{We appreciate that in clinical practice you would need to read the full article to make a final decision. However, the full article is not always freely available and people sometimes make decisions based on the abstract.}\\
\texttt{Based on the information reported in the abstract, please answer the following question about the treatment. Only output a score between 0-10. Do not provide any explanations.}\\

\texttt{Abstract: \textbf{\{ABSTRACT\}}}\\
\texttt{Question: \textbf{\{QUESTION\}}}
\end{tcolorbox}

For interpreting the trial results from the simplified abstracts, we used the following prompt:

\begin{tcolorbox}[colback=gray!10, colframe=gray!50, arc=4mm, breakable, enhanced, boxrule=0.4mm, toptitle=0mm, bottomtitle=0mm]
\small
\texttt{This summary describes a study evaluating a treatment in the field of cancer published in a peer-reviewed journal.}\\
\texttt{We appreciate that in clinical practice you would need to read the full article to make a final decision. However, the full article is not always freely available and people sometimes make decisions based on the summary.}\\
\texttt{Based on the information reported in the summary, please answer the following question about the treatment. Only output a score between 0-10. Do not provide any explanations.}\\

\texttt{Abstract: \textbf{\{ABSTRACT\}}}\\
\texttt{Question: \textbf{\{QUESTION\}}}
\end{tcolorbox}

\subsection{Reducing the Effect of Spin}
\label{appendix:reduce_effect_prompts}

We experimented with indicating whether the given abstract contains spin or not to see if this would help reduce LLMs’ tendency to overstate favorable trial results in the presence of spin. Based on either the reference label or model-inferred label, we determine the text that is inserted for \texttt{\{IF\_CONTAINS\}} in the prompt. If the label is ``spin'' or ``yes'', the text would be \texttt{contains}. For labels that are ``no\_spin'' or ``no'', we insert the text \texttt{does not contain}.

\begin{tcolorbox}[colback=gray!10, colframe=gray!50, arc=4mm, breakable, enhanced, boxrule=0.4mm, toptitle=0mm, bottomtitle=0mm]
\small
\texttt{This abstract describes a study evaluating a treatment in the field of cancer published in a peer-reviewed journal.}\\
\texttt{We appreciate that in clinical practice you would need to read the full article to make a final decision. However, the full article is not always freely available and people sometimes make decisions based on the abstract.}\\
\texttt{Spin or misrepresentation of study findings can be used to influence, positively, the interpretation of statistically nonsignificant randomized controlled trials (RCTs), for example, by emphasizing the apparent benefit of a secondary outcome or findings from a subgroup of patients.}\\
\texttt{If an abstract contains spin, the content should be treated with a higher level of skepticism.}\\
\texttt{Based on the information reported in the abstract and the fact that it \{IF\_CONTAINS\} spin, please answer the following question about the treatment. Only output a score between 0-10. Do not provide any explanations.}\\

\texttt{Abstract: \textbf{\{ABSTRACT\}}}\\
\texttt{Question: \textbf{\{QUESTION\}}}
\end{tcolorbox}

Below is the prompt that we used for the ``joint prompting'' (spin detection + result interpretation) strategy. We prompt each question individually for a given abstract.

\begin{tcolorbox}[colback=gray!10, colframe=gray!50, arc=4mm, breakable, enhanced, boxrule=0.4mm, toptitle=0mm, bottomtitle=0mm]
\small
\texttt{This abstract describes a study evaluating a treatment in the field of cancer published in a peer-reviewed journal.}\\
\texttt{We appreciate that in clinical practice you would need to read the full article to make a final decision. However, the full article is not always freely available and people sometimes make decisions based on the abstract.}\\
\texttt{Spin or misrepresentation of study findings can be used to influence, positively, the interpretation of statistically nonsignificant randomized controlled trials (RCTs), for example, by emphasizing the apparent benefit of a secondary outcome or findings from a subgroup of patients.}\\
\texttt{Answer the following two questions with one JSON that contains the following keys:}\\
\texttt{detection: Does the following clinical trial abstract contain spin (yes/no)? Answer only with `yes' or `no'. Do not provide any explanations.}\\
\texttt{interpretation: Based on this abstract and previous spin detection, \textbf{\{QUESTION\}} Only output a score between 0-10. Do not provide any explanations. If an abstract contains spin, the contents should be treated with a higher level of skepticism.}\\

\texttt{Abstract: \textbf{\{ABSTRACT\}}}
\end{tcolorbox}

\subsection{Simplification of Abstracts} 
\label{appendix:simplification_prompt}

For simplifying medical abstracts for laypeople, we used the prompt used by \citet{joseph2024factpico}.

\begin{tcolorbox}[colback=gray!10, colframe=gray!50, arc=4mm]
\small
\texttt{My fifth grader asked me what this passage means: \textbf{\{ABSTRACT\}}}\\
\texttt{Help me summarize it for him, in plain language a fifth grader can understand.}
    
\end{tcolorbox}

\section{In-Depth Interpretation Results for Select Models}
\label{appendix:select_models}

Figure~\ref{fig:top_6_models_interpretation} shows the mean differences for the five interpretation questions from six LLMs with the best spin detection performance, comparing with human expert interpretations. The six models achieved a spin detection accuracy greater than 0.80, as reported in Section~\ref{section:results}. Although these models demonstrated strong performance in correctly detecting spin, they were generally more susceptible to spin in abstracts than human experts.

\begin{figure*}
\floatconts
  {fig:top_6_models_interpretation}
  {\caption{Mean differences of all five interpretation questions from top 6 LLMs in spin detection accuracy compared to human experts. Error bars represent 95\% confidence intervals. Positive mean differences indicate that LLMs interpreted spun abstracts as showing more favorable treatment results, while negative mean differences suggest that unspun abstracts were perceived as more favorable. This plot highlights that, despite strong spin detection performance, LLMs tend to overestimate the differences between spun and unspun abstracts relative to human experts.}}
  {\includegraphics[width=\linewidth]{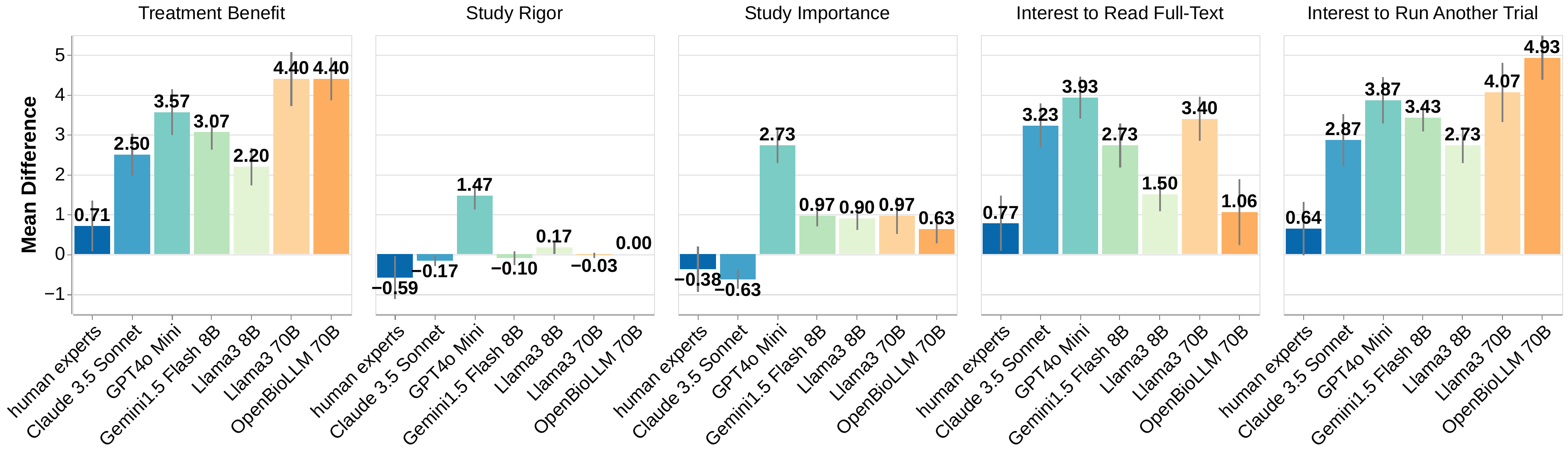}}
\end{figure*}

\section{Linear Regression Models}
\label{appendix:linear_models}

Figures~\ref{fig:spin_label_regression_rigor}, \ref{fig:spin_label_regression_importance}, \ref{fig:spin_label_regression_full_text}, and \ref{fig:spin_label_regression_another_trial} show the coefficients and 95\% confidence intervals from the linear regression models run with
an indicator for the presence of spin in an abstract as a predictor for study rigor, study importance, interest to read the full-text, and interest to run another trial.

\begin{figure*}
\floatconts
  {fig:spin_label_regression_rigor}
  {\caption{Coefficients from linear regression models with 95\% CI for each LLM showing how much different LLMs overestimate the rigor of study, when abstracts contain `spin'. In comparison with human experts (-0.59), LLMs show slightly greater susceptibility to spin.}}
  {\includegraphics[width=0.8\linewidth]{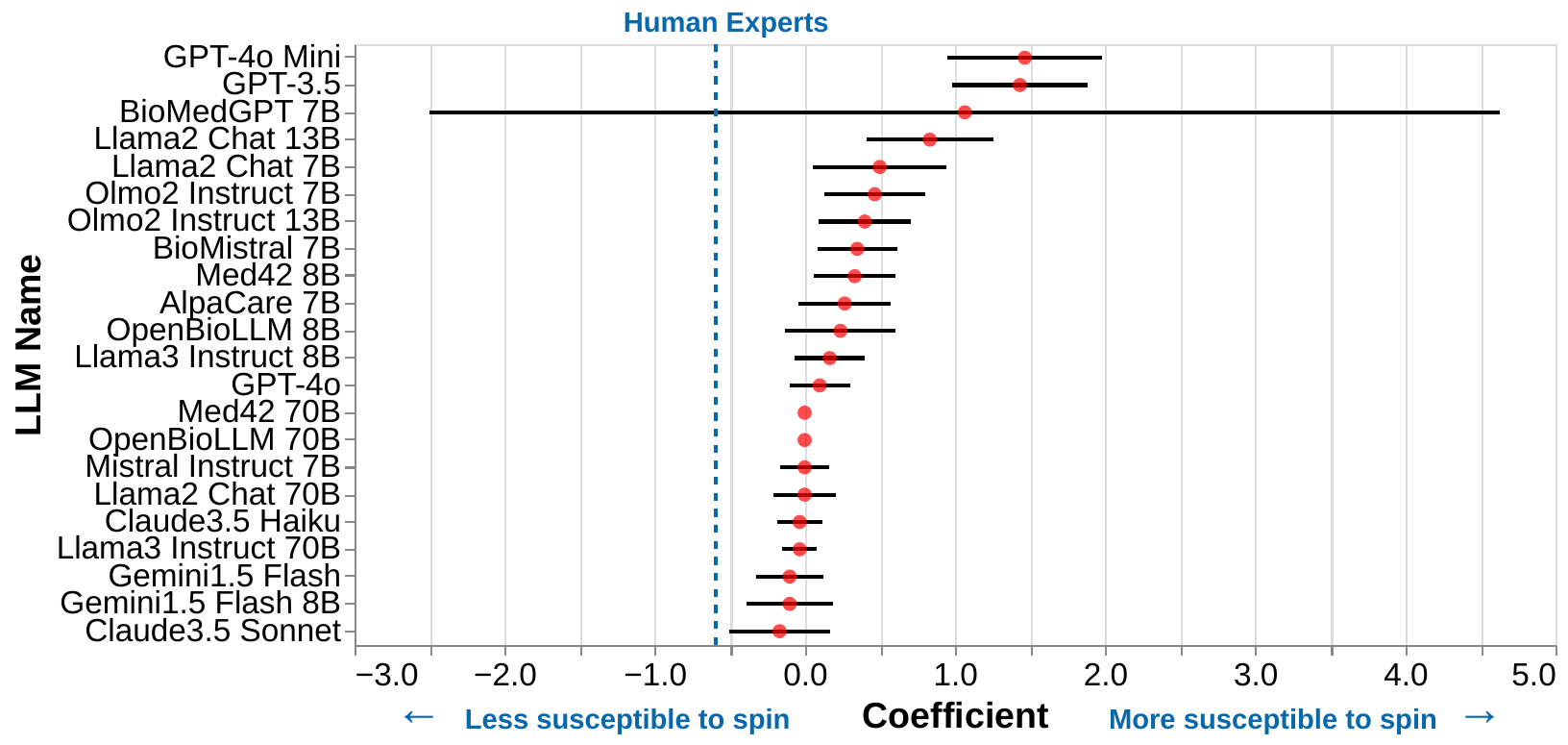}}
\end{figure*}

\begin{figure*}
\floatconts
  {fig:spin_label_regression_importance}
  {\caption{Coefficients from linear regression models with 95\% CI for each LLM showing how much different LLMs overestimate the importance of study, when abstracts contain `spin'. In comparison with human experts (-0.38), most LLMs show greater susceptibility to spin.}}
  {\includegraphics[width=0.8\linewidth]{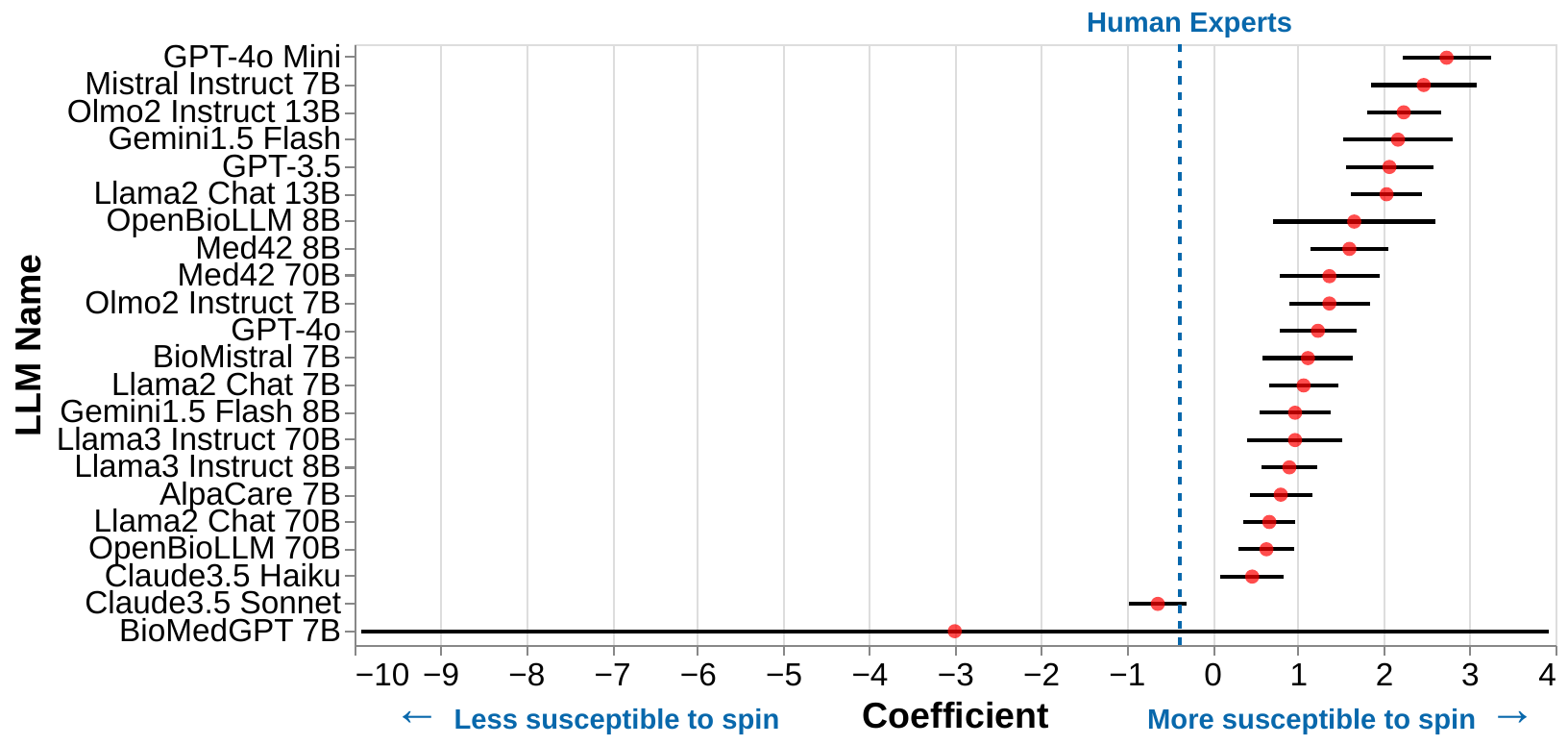}}
\end{figure*}

\begin{figure*}
\floatconts
  {fig:spin_label_regression_full_text}
  {\caption{Coefficients from linear regression models with 95\% CI for each LLM showing how much different LLMs overestimate the interest in full-text, when abstracts contain `spin'. In comparison with human experts (0.77), most LLMs show greater susceptibility to spin.}}
  {\includegraphics[width=0.8\linewidth]{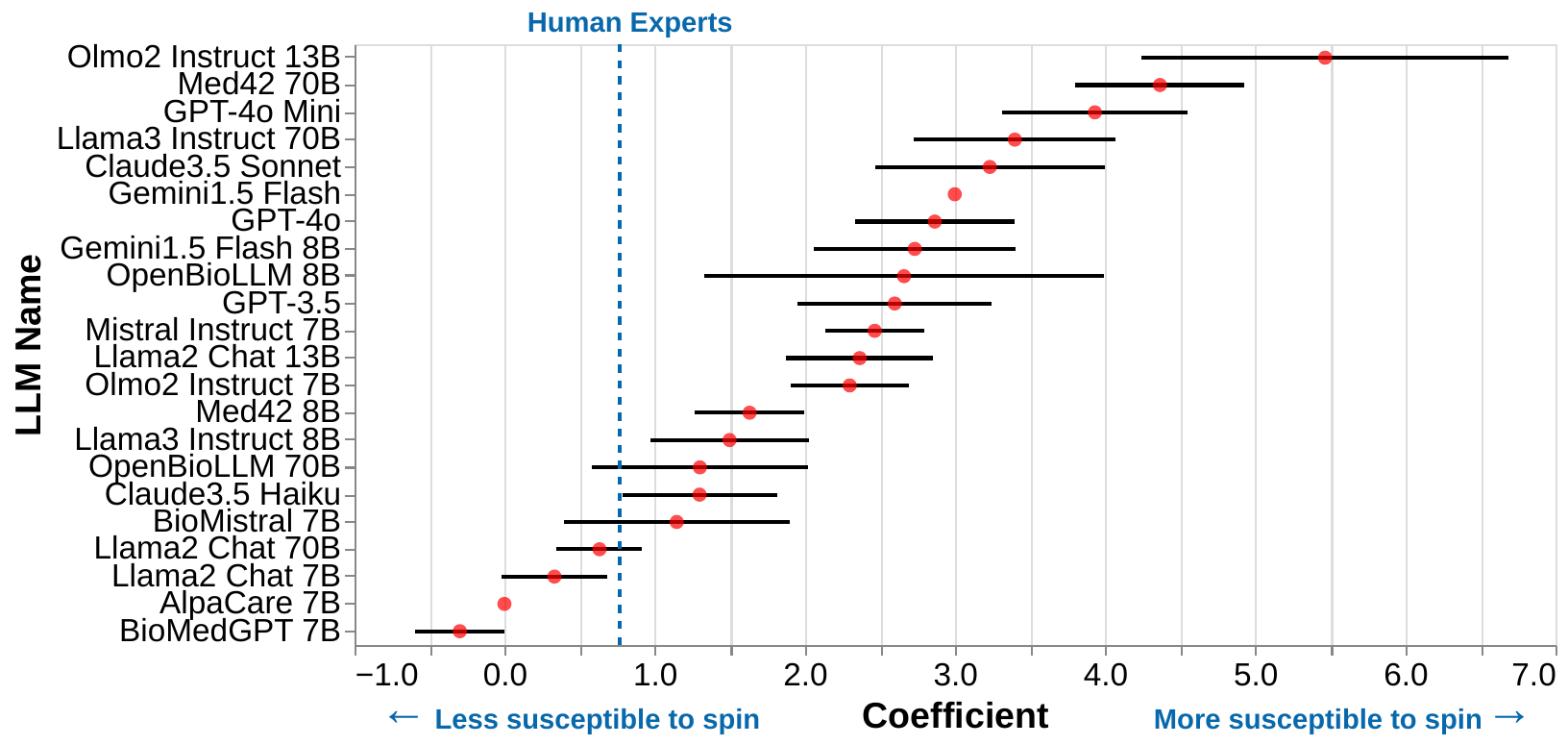}}
\end{figure*}

\begin{figure*}
\floatconts
  {fig:spin_label_regression_another_trial}
  {\caption{Coefficients from linear regression models with 95\% CI for each LLM showing how much different LLMs overestimate the interest in another trial, when abstracts contain `spin'. In comparison with human experts (0.64), most LLMs show greater susceptibility to spin.}}
  {\includegraphics[width=0.8\linewidth]{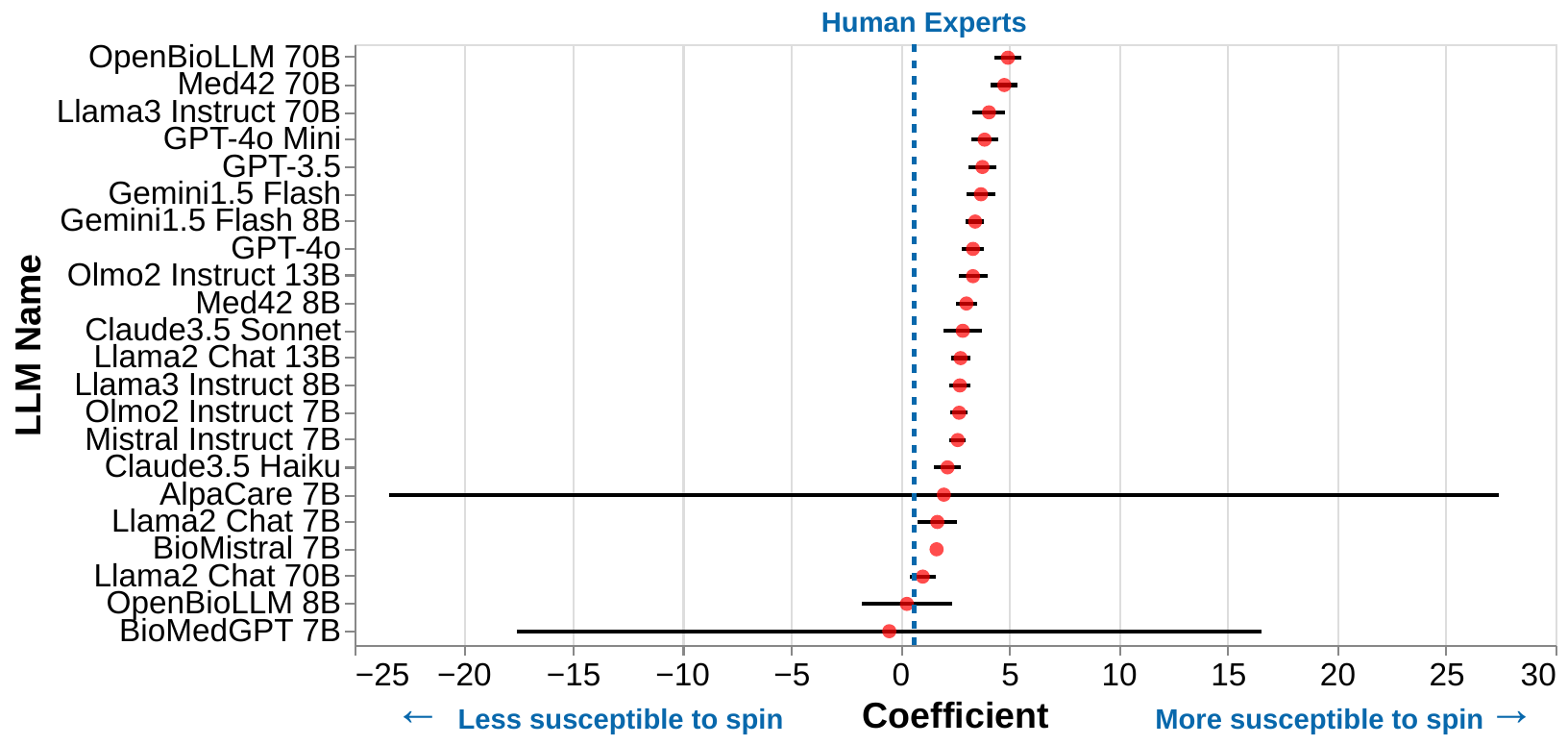}}
\end{figure*}

\section{Simplified Abstract Examples}
\label{appendix:simplification_examples}

Table~\ref{tab:example_pls} shows selected examples of simplified abstracts generated by LLMs. Further examples can be found in our project GitHub repo.

\begin{table*}
\footnotesize
\floatconts
  {tab:example_pls}
  {\caption{Examples of simplified abstracts generated by LLMs.}}
  {\begin{tabular}{p{0.97\textwidth}}
  \toprule
  \rowcolor{gray!20}
  \textbf{Model: \textcolor{RoyalBlue}{OpenBioLLM 70B} --- Original abstract type: \textcolor{red}{spin}} \newline \textbf{Plain language version}: This study looked at whether a certain treatment (called treatment A) could help people with brain tumors live longer and stay independent after surgery or radiosurgery. The researchers studied 359 patients who had one to three brain tumors and gave them either treatment A or just observed them without treatment. They found that treatment A did not make a difference in how long patients stayed independent or how long they lived overall. However, it did reduce the chance of the tumors coming back in the brain, which led to fewer deaths from the tumors spreading in the brain. So, while treatment A didn't improve independence or overall survival, it did help lower the risk of the tumors returning and causing problems in the brain.\\
  \textbf{Model: \textcolor{RoyalBlue}{OpenBioLLM 70B} --- Original abstract type: \textcolor{teal}{no spin}} \newline \textbf{Plain language version}: This study looked at whether a certain treatment (called treatment A) could help people with cancer in their brains live longer and stay independent after surgery or radiosurgery. The researchers compared a group of patients who received treatment A to a group of patients who were just observed and didn't receive any extra treatment. They found that there was no big difference between the two groups in terms of how long they lived or how long they were able to do their daily activities. However, they did find that the group of patients who received treatment A had less progression of the cancer in their brains compared to the group that was just observed. There were also some side effects reported, but overall, treatment A was not found to be better than just observing the patients for most outcomes.\\ \midrule
  \rowcolor{gray!20}
  \textbf{Model: \textcolor{RoyalBlue}{Gemini 1.5 Flash 8B} --- Original abstract type: \textcolor{red}{spin}} \newline \textbf{Plain language version}: Doctors are doing a big study to see if a new medicine (Treatment A) helps kids with a type of blood cancer called AML get better.  They looked at a bunch of kids getting the standard treatment, and another group getting the standard treatment *plus* Treatment A.  The kids in both groups were pretty much the same. The study found that Treatment A helped the kids' blood counts get better and stay better longer.  But, overall, the medicine didn't really change how long the kids lived. The survival rate was about the same for both groups (25\% to 32\% after 3 years). However, there was one important difference:  Girls who got Treatment A did better than boys.  Also, kids who only needed one round of the standard treatment did better.  The study also found that using Treatment A itself was a good thing for survival. The doctors need to do more studies to see if the girls did better because of their sex, or if there's something else going on with their blood cancer.  They need to look at the cancer's chromosomes to see if that plays a role.\\
  \textbf{Model: \textcolor{RoyalBlue}{Gemini 1.5 Flash 8B} --- Original abstract type: \textcolor{teal}{no spin}} \newline \textbf{Plain language version}: Doctors are doing a big study to see if a new medicine (Treatment A) helps kids with a type of blood cancer called acute myeloid leukemia (AML). They gave Treatment A to some kids and a regular medicine (cytarabine and idarubicine) to other kids.  They looked at how well the kids got better and how long they lived. The kids who got Treatment A did about the same as the kids who got the regular medicine.  The percentage of kids who got better quickly (complete response) was almost the same in both groups.  Also, the number of kids who lived for 3 years was not much different between the groups. So, the study showed that Treatment A wasn't better than the regular medicine at helping kids with AML.\\ \midrule
  \rowcolor{gray!20}
  \textbf{Model: \textcolor{RoyalBlue}{Llama3 Instruct 8B} --- Original abstract type: \textcolor{red}{spin}} \newline \textbf{Plain language version}: So, this passage is talking about a study that was done to help people who are about to start chemotherapy (a type of cancer treatment). The study wanted to see if giving them some information and education before they started chemotherapy would help them feel less stressed and worried. The study found that giving people this information did help them feel less worried about some specific things, like getting sick to their stomach or feeling uncomfortable during the treatment. But it didn't really help them feel less stressed overall. The study also found that the information helped some people who were already feeling really stressed and worried, but it didn't help everyone. The researchers think that this information might be helpful for some people, but they need to do more research to make sure it will work for people who are different from the ones in this study. So, in simple terms, the study was trying to figure out if giving people information before chemotherapy would help them feel better, and it found that it did help with some specific things, but not with feeling stressed overall.\\
  \textbf{Model: \textcolor{RoyalBlue}{Llama3 Instruct 8B} --- Original abstract type: \textcolor{teal}{no spin}} \newline \textbf{Plain language version}: So, the passage is talking about a study that was done to see if giving patients information about what to expect before they start chemotherapy would help reduce their stress and anxiety. Imagine you're going to get a shot, and you don't know what it's going to feel like or what might happen. You might be worried or scared. That's kind of what it's like for people who are about to start chemotherapy. They don't know what it's going to feel like or what side effects they might get. The study was trying to see if giving patients some information beforehand would help them feel less stressed and anxious. They gave some patients this information, and then they compared them to patients who didn't get the information. But, it turned out that the patients who got the information didn't actually feel less stressed or anxious than the patients who didn't get the information. The only thing that did change was that the patients who got the information were a little less worried about some specific things, like how their body would feel during the treatment. But, overall, the study found that giving patients information beforehand didn't really help them feel better. So, the conclusion is that just giving patients information before chemotherapy isn't enough to really help them feel less stressed or anxious.\\
  \bottomrule
  \end{tabular}}
\end{table*}

\section{Detailed Results from Reducing the Effect of Spin}
\label{appendix:spin_reduction}

Table~\ref{tab:spin_reduction_strategy_results} presents mean differences across different spin mitigation approaches for all LLMs, compared to the ``baseline.'' Figure~\ref{fig:baseline_detect_interpret_diff} shows how ``detect + interpret''(prompting the model to first detect if given abstract contains spin or not and then to answer the interpretation question based on its answer for spin detection) reduces all LLMs' tendency to overstate favorable trial results compared to ``baseline.'' This shows that the strategy works even when the model is not a strong spin detector.

\begin{table*}
\footnotesize
\floatconts
  {tab:spin_reduction_strategy_results}
  {\caption{This table presents the results of using different spin mitigation strategies for all 22 LLMs. Our results show that reducing the effects of spin in interpretations can be achieved by careful prompting.}}
  {\begin{tabular}{ccccc}
  \toprule
\multicolumn{1}{c}{\textbf{Model}} &
  \multicolumn{1}{c}{\textbf{Baseline}} &
  \multicolumn{1}{c}{\textbf{+ Ref Label}} &
  \multicolumn{1}{c}{\textbf{+ Model Label}} &
  \multicolumn{1}{c}{\textbf{Detect + Interpret}} \\ \midrule
\rowcolor{gray!20} AlpaCare 7B         & 6.05     & 1.6        & 4.33          & -1.29              \\
BioMedGPT 7B        & 1.24     & 0.00       & 0.00          & 0.98               \\
\rowcolor{gray!20} BioMistral 7B       & 1.67     & 0.23       & 1.72          & 1.00               \\
Claude3.5 Haiku     & 2.97     & 2.47       & 2.23          & 0.73               \\
\rowcolor{gray!20} Claude3.5 Sonnet    & 2.50     & 1.27       & 1.63          & 0.87               \\
Gemini1.5 Flash     & 2.50     & 1.07       & 1.23          & 0.10               \\
\rowcolor{gray!20} Gemini1.5 Flash 8B  & 3.07     & 0.70       & 1.83          & 0.90               \\
GPT-3.5              & 3.90     & 2.93       & 3.80          & 2.53               \\
\rowcolor{gray!20} GPT-4o               & 3.13     & 1.27       & 1.93          & 0.90               \\
GPT-4o Mini          & 3.57     & 1.80       & 2.27          & 1.33               \\
\rowcolor{gray!20} Llama2 Chat 13B     & 2.93     & 1.00       & 1.10          & 0.10               \\
Llama2 Chat 70B     & 2.69     & 1.84       & 2.00          & 1.53               \\
\rowcolor{gray!20} Llama2 Chat 7B      & 3.50     & 0.83       & 2.30          & 1.67               \\
Llama3 Instruct 70B & 4.40     & 3.97       & 4.10          & 2.10               \\
\rowcolor{gray!20} Llama3 Instruct 8B  & 2.20     & 1.60       & 1.83          & 0.07               \\
Med42 70B           & 4.83     & 3.70       & 4.50          & 3.60               \\
\rowcolor{gray!20} Med42 8B            & 2.97     & 1.47       & 2.13          & 1.07               \\
Mistral Instruct 7B & 3.90     & 2.60       & 2.87          & 1.40               \\
\rowcolor{gray!20} Olmo2 Instruct 13B  & 6.00     & 3.47       & 4.63          & 1.07               \\
Olmo2 Instruct 7B   & 3.23     & 3.10       & 3.36          & 1.50               \\
\rowcolor{gray!20} OpenBioLLM 70B      & 4.40     & 2.93       & 3.83          & 3.03               \\
OpenBioLLM 8B       & 2.74     & 2.83       & 3.30          & -0.11              \\ \bottomrule
\end{tabular}}
\end{table*}

\begin{figure*}
\floatconts
  {fig:baseline_detect_interpret_diff}
  {\caption{Mean differences for the treatment benefit question between the ``baseline'' and ``detect + interpret'' approaches for each LLM. The ``baseline'' score is shown in \textbf{black}, while the ``detect + interpret'' score is in \textcolor{orange}{\textbf{orange}}. LLMs are ordered from top to bottom based on their spin detection performance, with the best-performing model at the top and the worst at the bottom.  Regardless of the original spin detection accuracy, the ``detect + interpret'' approach consistently reduces the effect of spin on treatment benefit assessments across all LLMs.}}
  {\includegraphics[width=\linewidth]{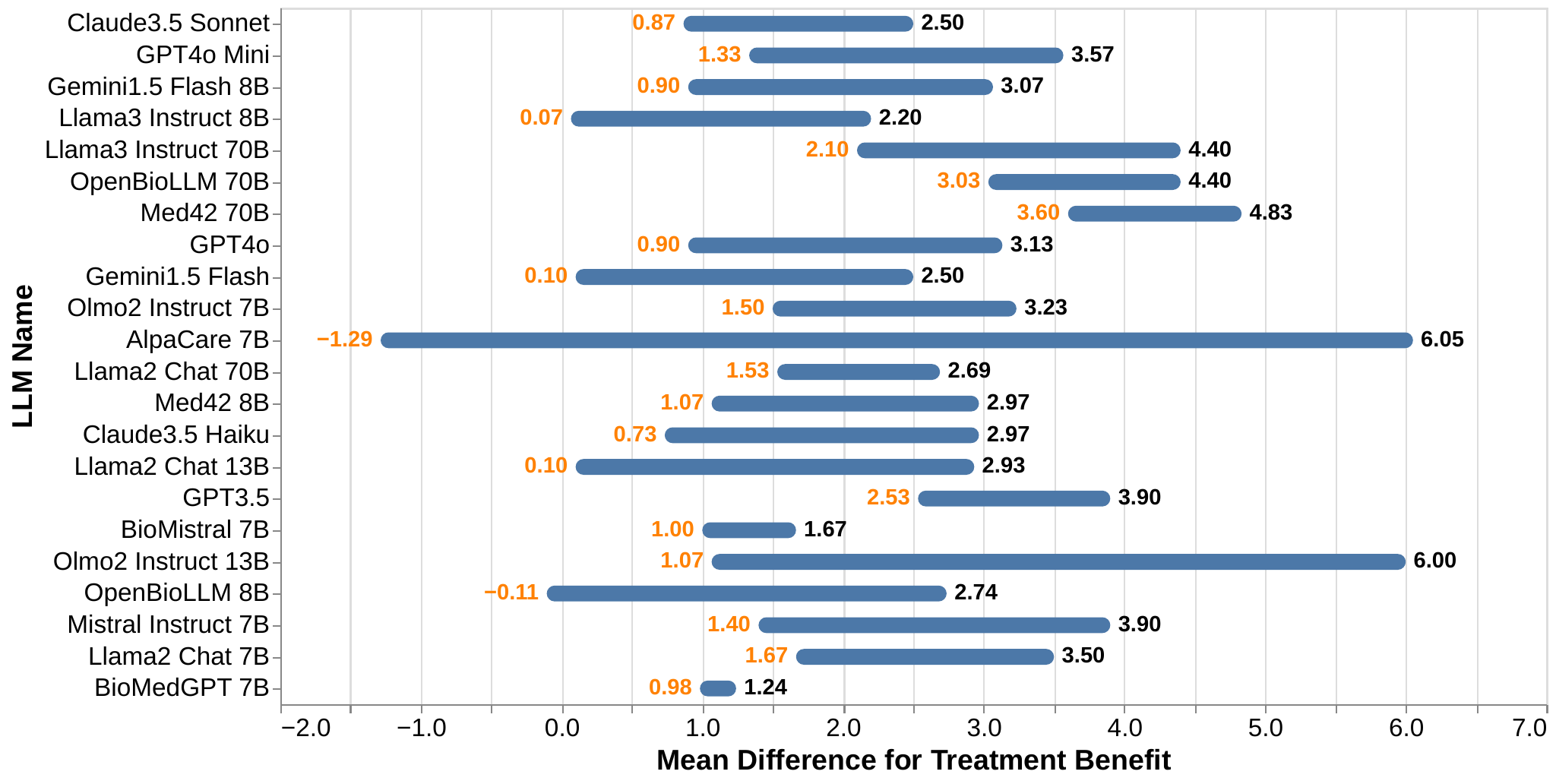}}
\end{figure*}

\end{document}